\documentclass{article}

\PassOptionsToPackage{numbers}{natbib}
\usepackage[main,final]{neurips_2026}

\usepackage[utf8]{inputenc}
\usepackage{hyperref}
\usepackage{url}
\usepackage{booktabs}
\usepackage{amsfonts}
\usepackage{amsmath}
\usepackage{amssymb}
\usepackage{microtype}
\usepackage{xcolor}
\usepackage{graphicx}
\usepackage{enumitem}
\usepackage{wrapfig}
\usepackage{algorithm}
\usepackage{algpseudocode}

\usepackage[capitalize]{cleveref}

\usepackage{amsthm}

\Crefname{equation}{Eq.}{Eqs.}
\Crefname{figure}{Fig.}{Figs.}
\Crefname{tabular}{Tab.}{Tabs.}
\Crefname{appendix}{App.}{Apps.}

\IfFileExists{todonotes.sty}{\usepackage[disable]{todonotes}}{%
  \newcommand{\todo}[2][]{}%
  \newcommand{\listoftodos}[1][]{}%
}

\newcommand{\MAEEloAvg}{17.9}

\newcommand{\Elo}{\mathrm{Elo}}
\newcommand{\eloLLM}{\Elo_{\mathrm{LLM}}}
\newcommand{\eloHuman}{\Elo_{\mathrm{Human}}}
\newcommand{\eloLLMi}[1]{\Elo_{\mathrm{LLM},#1}}
\newcommand{\eloHumani}[1]{\Elo_{\mathrm{Human},#1}}

\newcommand{\bstar}{\beta^{*}}

\graphicspath{{figures/}}

\title{From Uncertain Judgments to Calibrated Rankings: Conformal Elo Estimation for LLM Evaluation}

\author{Bora Kargi \\
    ELLIS Institute Tübingen \\
    OpenEuroLLM \\
\And David Salinas \\
    ELLIS Institute Tübingen \\
    OpenEuroLLM
}

\begin{document}

\maketitle

\vspace{-12pt}
\begin{center}
\small
\href{https://github.com/kargibora/SoftElo}{%
  \raisebox{-0.2em}{\includegraphics[height=1.2em]{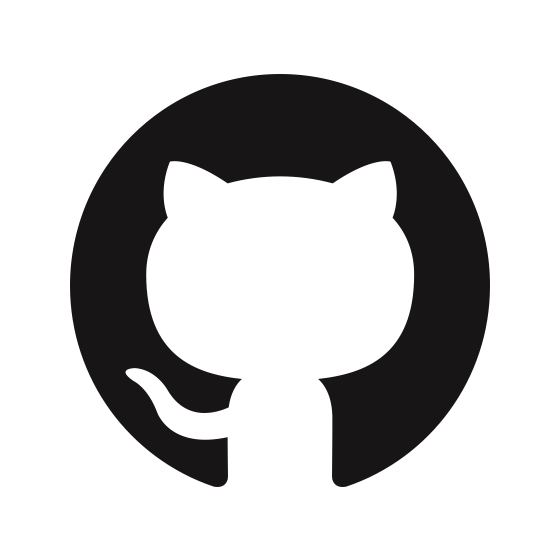}}\enspace Code}%
\qquad
\href{https://huggingface.co/datasets/kbora/SoftElo}{%
  \raisebox{-0.2em}{\includegraphics[height=1.2em]{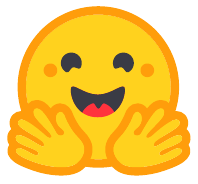}}\enspace Dataset}
\end{center}
\vspace{12pt}

\begin{abstract}
Evaluating new large language models typically requires costly human annotation campaigns at scale. LLM-as-a-judge offers a cheaper alternative, but judge scores carry systematic errors — such as position bias, self-preference, or intransitivity — that can strongly miscalibrate the resulting rankings. We quantify the resulting judge--human disagreement at two complementary levels. At the local level, we estimate per-battle uncertainty from the judge's own score differences by propagating calibrated win probabilities rather than hard labels into the Bradley--Terry procedure. This alone provides a drastic improvement to Elo estimation accuracy, bringing LLM-derived ratings within $\MAEEloAvg{}$ Elo MAE of human-derived ones when averaged over 55 held-out models on LMArena. At the global level, we apply split conformal prediction to the residual gap between LLM-derived and human-derived Elo ratings across held-out models, producing prediction intervals with distribution-free marginal coverage guarantees that account for irreducible LLM--human disagreement. Together, these two layers yield a low-cost evaluation tool that provides developers with calibrated Elo estimates and honest uncertainty bounds, without access to large-scale human annotations.

\end{abstract}

\section{Introduction}
\label{sec:introduction}

Modern LLM evaluation rests on pairwise comparison: platforms like Chatbot
Arena~\citep{chiang2024chatbot} or ComparIA \cite{termignon2026comparia}
elicit human votes between two completions and aggregate them
into a Bradley--Terry (BT) leaderboard. 
LLM-as-a-judge evaluation~\citep{zheng2023judging, dubois2024length, li2024arenahard}
offers a cheaper proxy --- a judge model produces the verdicts --- and is now the dominant
mode of leaderboard construction when side-stepping the cost of human annotations is required.

Despite their cost advantage and support for fully automated evaluation, LLM judges suffer from well-documented biases, including: a position bias toward the first completion~\cite{ko2024judging}; a length bias favoring longer answers~\cite{dubois2024length}; and a self-preference bias toward completions generated by — or resembling — the judge itself~\cite{panickssery2024}. While some of those issues can be addressed easily (for instance the positional bias can be addressed by randomizing the position or by averaging both positions), the last is particularly damaging since most leaderboards rely on closed-weight judges~\cite{salinastuning}, which can be deprecated or silently updated at any time, rendering prior evaluations irreproducible. A win rate against a fixed baseline is conditional on that opponent: non-transitive preferences can change the ranking when the baseline changes, while an overly strong or weak baseline produces nearly one-sided outcomes that poorly distinguish candidates \cite{nontransitivity2025,donyehiya2026mediocrity}.

These issues are not merely theoretical. On Arena-Hard, the reported win rate of Gemini-2.5
shifts from $79.0\% \pm 2\%$ to $49.1\% \pm 2.5\%$ depending on the judge, moving the model from 2nd to 8th place\footnote{See the 
\href{https://github.com/lmarena/arena-hard-auto/blob/196f6b8/README.md}{Arena-Hard leaderboard}, accessed May 7, 2026.}. This illustrates how strongly the choice of judge can reshape a leaderboard: the reported uncertainty around each win rate is small relative to the shift induced by changing the judge.
Motivated by this fragility, we study an automated benchmarking protocol that
estimates confidence intervals for Elo ratings directly from LLM-judged battles across multiple opponents, rather than reporting win rate against a fixed baseline. This targets the quantity leaderboard users ultimately inspect: distances on the human Elo scale. Concretely, each model is evaluated through battles against multiple opponents, and the resulting comparisons are aggregated with Bradley--Terry.

To provide accurate and reliable uncertainty estimates for Elo ratings, we improve the uncertainty handling at two complementary scales.
At a \textit{local} scale --- the scale of individual instructions --- we show that modeling judge uncertainty is critical: LLM judge scores are systematically biased toward stronger models. To alleviate this, we estimate calibrated win probabilities from judge scores via maximum likelihood, which drastically reduces estimation error.

At the \textit{global} scale, even with calibrated local probabilities, a residual gap remains between LLM-derived and human-derived Elo ratings. To account for this irreducible discrepancy and quantify the uncertainty at this level, we apply split conformal prediction to the LLM--human
Elo residuals across held-out models. This calibration operates at the level of aggregate Elo estimates --- the quantity practitioners actually report --- rather than individual pairwise judgments, producing conformal intervals with distribution-free marginal coverage guarantees over new models drawn from the same population. Together, these two layers yield a low-cost evaluation framework that provides developers with accurate Elo estimates and honest uncertainty bounds without requiring human annotations for the model under test.

\begin{figure}[!t]
\centering
\includegraphics[width=\linewidth]{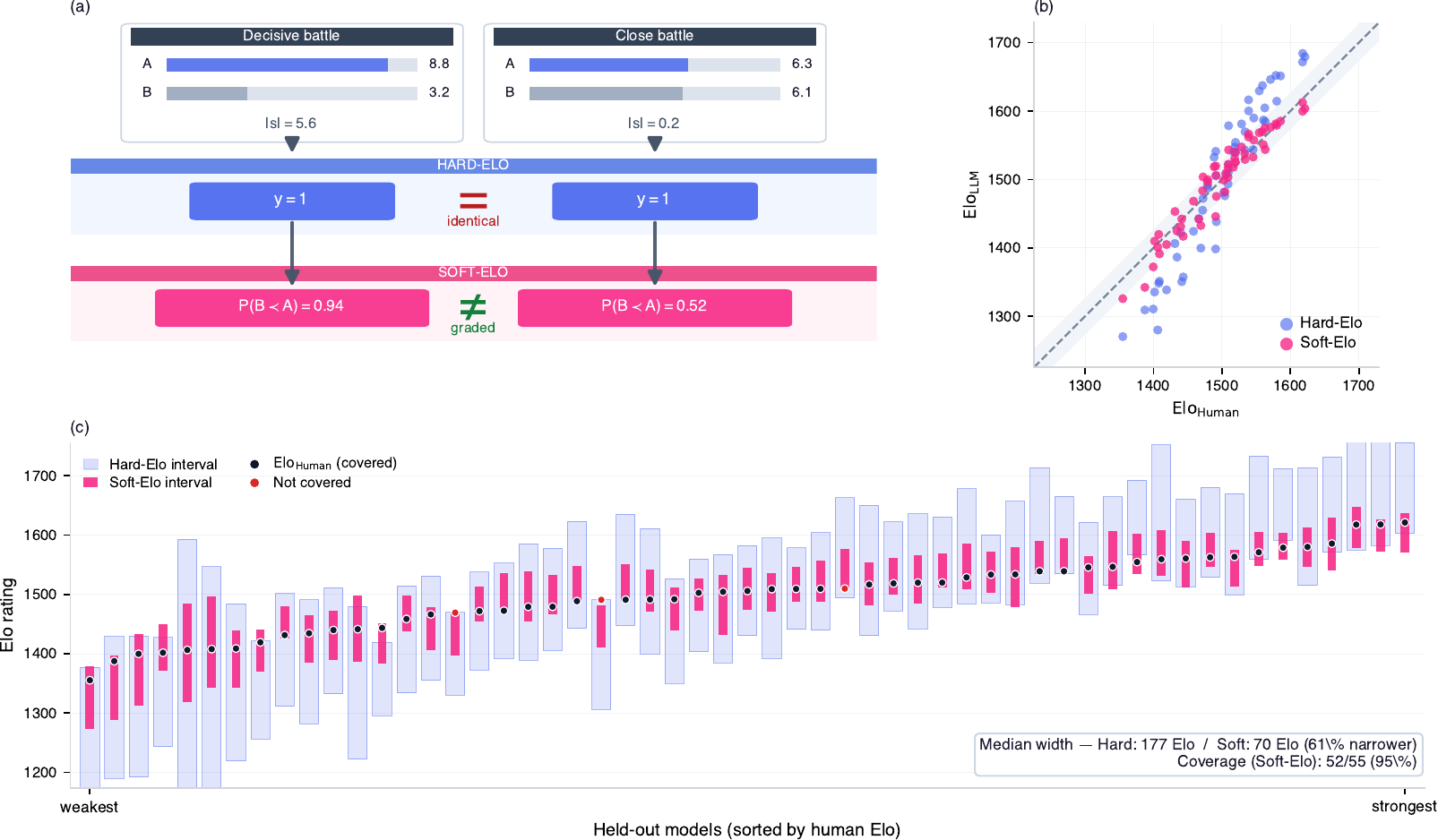}
\caption{\textbf{Calibrated win probabilities sharpen Elo estimates and yield narrow conformal intervals.}
Hard labels in blue, calibrated targets in pink throughout.
(a)~Hard labels collapse each battle into $\{0, 0.5, 1\}$; the calibrated target $P[B \prec A \mid x]$ preserves the per-battle score difference.
(b)~Held-out Elo for Qwen3.5-27B: hard-label fits fan from the diagonal; calibrated fits cluster on it.
(c)~Leave-one-model-out conformal intervals at $90\%$ coverage over $55$ models; calibrated intervals are markedly narrower at matched coverage.}
\label{fig:teaser}
\end{figure}

We summarize our contributions as follows:
\begin{itemize}
\item We propose an LLM benchmark that estimates Elo ratings from battles against diverse opponents, avoiding the fragility of fixed-baseline comparisons while quantifying uncertainty at both local (per-judgment) and global (per-model) levels.
\item At the local level, regressing calibrated win probabilities from raw judge scores — rather than reducing them to binary labels — yields well-calibrated per-instruction uncertainty and drastically improves Elo estimation accuracy.
\item At the global level, we apply split conformal prediction to LLM--human Elo residuals across held-out models, producing conformal intervals with distribution-free marginal coverage guarantees over new models.
\item On LMArena, our pipeline estimates Elo ratings with mean absolute error $\MAEEloAvg{}$ Elo on held-out models compared to human ratings, offering a low-cost alternative to large-scale human annotation campaigns.
\end{itemize}

\section{Related Work}
\label{sec:related-work}

\paragraph{LLM-as-a-judge benchmarks.}

Pairwise LLM-as-a-judge benchmarks have become the default
cheap surrogate for human preference rankings, an alternative to crowd-sourced platforms \cite{chiang2024chatbot,termignon2026comparia}. 
AlpacaEval \cite{dubois2024length}, Arena-Hard \cite{li2024arenahard}, and MT-Bench \cite{zheng2023judging} all score a new model against a fixed reference using GPT-4 as judge, with MT-Bench considering multiturn.
While most benchmarks report win rates against a fixed baseline — an approach shown to be fragile under non-transitive judge preferences \cite{nontransitivity2025} — some methods instead fit Bradley-Terry models to LLM judge annotations across multiple opponents. Previous work has automated the full pipeline from question generation to committee judging \cite{zhao2024autoarena} or used prediction-powered inference to debias BT coefficients by combining judge labels with a small set of human annotations \cite{boyeau2024autoeval}. Our work follows a similar direction but requires no human labels for the evaluated model, instead handling uncertainty through calibrated local probabilities and conformal prediction at the global level.
Soft-Elo also applies to committees: Appendix~\ref{app:committees} compares
categorical, vote-share, and calibrated-score aggregation on shared battles.

\paragraph{Judge biases.}
Systematic biases for LLM-as-a-judge benchmarks with fixed baselines are well documented at the local level, e.g. at the level of individual battles:
position bias~\citep{ko2024judging, wang2024large}, verbosity
bias~\citep{dubois2024length, saito2024verbosity}, 
self-preference~\citep{panickssery2024, koo2024selfpreference} or the problematic sensitivity to the baseline chosen \cite{nontransitivity2025,donyehiya2026mediocrity}. Biases at the global level, e.g. at the leaderboard level have also been studied including for instance the sensitivity to which comparisons are
observed~\citep{droppingpreferences2025} or the benchmark composition
effects that can favor proprietary models~\citep{leaderboardillusion2025}.
Closest in spirit to our work is Bridge~\citep{bridge2025}, which explicitly models systematic judge--human discrepancies under both absolute and pairwise evaluation. We instead keep the ranking model fixed and change the targets fed into it.

\paragraph{Bradley--Terry fitting for LLM leaderboards.}
LLM leaderboards built on top of human annotations usually fit Bradley--Terry to hard win/tie/loss labels
and report the resulting Elo scores~\citep{chiang2024chatbot,
zheng2023judging, boubdir2023elo}. Recent extensions mostly modify the
ranking model rather than the labels, including tie- and
covariance-aware arena models~\citep{statframework2024}, stable or
annotator-aware Elo variants~\citep{amelo2025}, methods for
intransitive preferences~\citep{gpm2024}, and non-parametric
alternatives~\citep{nonparametriceval2026}. Adjacent reward-modeling
work makes a similar point for ties and ordinal feedback: binary labels
discard useful signal that can be preserved in richer
targets~\citep{rewardties2024, afsharrad2026ordinal, liu2024ordinal}.
Departing from previous work, we leave the BT model unchanged and modify only its \emph{training targets}, to account for the uncertainty estimated at the instruction level.

\paragraph{Judgment distributions and soft preferences.}
The fact that hard single-point judgments or purely
ordinal feedback throw away useful information have been noted in recent work ~\citep{judgmentdistribution2025,
distributionalignment2025,
confidenceautoraters2025, whitfill2025cardinal}. Related calibration
methods include LLM-Rubric~\citep{hashemi2024llmrubric}, which predicts
human ratings from multidimensional rubric responses, and Quantitative
LLM Judges~\citep{sahoo2025quantitative}, which aligns judge scores to
human scores with post-hoc regression. Our method is complementary: we regress calibrated probabilities from LLM judgments and use them as soft targets for the BT model so that the aggregate Elo axis better matches the human scale.

\paragraph{Uncertainty quantification for LLM evaluation.}
Recent work applies conformal prediction to individual judge
scores~\citep{intervaleval2025} or uses selective conformal prediction
to abstain on unreliable pairwise calls~\citep{scope2026}. Prediction-powered
methods~\citep{predictionpowered2024, stratppi2024, park2025rautoeval}
provide a complementary route by combining limited human supervision
with abundant automatic labels to build ranking uncertainty sets. Our
target differs: we construct conformal intervals for a new model's Elo performance on the global leaderboard and show that interval widths are reduced significantly by forwarding the uncertainty information at the local battle level. 
A broader survey of conformal methods for NLP is given by~\cite{cpcpnlpsurvey2024}.

\section{Background}
\label{sec:background}

\subsection{Bradley--Terry Model and Elo Ratings}
\label{sec:bt-elo}

Following standard leaderboard practice, we fit a BT
model~\citep{bradley1952rank} to pairwise \textbf{battle} outcomes. A battle
compares two model completions $A$ and $B$ on the same instruction $x$,
with either an LLM or human judge producing a verdict. Each model $i$ carries a latent strength
$\theta_i \in \mathbb{R}$, and the probability that $i$ beats $j$ is
$\sigma(\theta_i - \theta_j)$ where $\sigma$ is the logistic. Strengths are
estimated by maximising the regularised log-likelihood of the observed
battle outcomes over the set $\mathcal{B}$:
\begin{equation}
  \hat{\boldsymbol{\theta}}
  \;=\;
  \arg\max_{\boldsymbol{\theta}}
  \sum_{(i,j,x)\,\in\,\mathcal{B}}
  \Bigl[
    y_{ij}(x)\log\sigma(\theta_i{-}\theta_j)
    + \bigl(1{-}y_{ij}(x)\bigr)\log\bigl(1{-}\sigma(\theta_i{-}\theta_j)\bigr)
  \Bigr]
  - \lambda\|\boldsymbol{\theta}\|^2,
  \label{eq:bt-mle}
\end{equation}
where $y_{ij}(x) \in \{0, 0.5, 1\}$ is the ternary battle label
($1$ if $A$ wins, $0$ if $B$ wins, $0.5$ for a tie) and
$\lambda = 0.01$. 
We follow the Chatbot Arena
convention~\citep{chiang2024chatbot} and split a tied battle into two
rows of weight $0.5$ (one per side); this is equivalent to
plugging $y = 0.5$ directly into Eq.~\ref{eq:bt-mle} and report results on the standard Elo scale
$\mathrm{Elo}_i = 1500 + (400/\ln 10)\,\theta_i$, an affine transform
of $\theta$ on which only differences are meaningful. We refer to this
pipeline as \textbf{Hard-Elo}.

\subsection{Judge Prompt}
\label{sec:judges}

We consider the instructions, pairwise completions, and accompanying human
preference labels $y^*_{ij}(x) \in \{0, 0.5, 1\}$ from LMArena 100K, a
multilingual subset of Chatbot Arena~\citep{chiang2024chatbot}. The corpus is
English-dominated (${\approx}58\%$ of instructions), with the remainder spanning
twelve further languages including Chinese, Japanese, Vietnamese, and German.

For each judge, we annotate approximately $25{,}000$ candidate battle rows,
sampled to spread model appearances across the $M=55$ models in the LMArena
100K subset. Since each battle contains two models, this corresponds to roughly
$900$ candidate model appearances per model on average.

The judge prompt used throughout this paper produces a scalar score
$s(x) \in \mathbb{R}$ for each battle $x$, where positive values indicate a
preference for completion $A$ and negative values favor completion $B$.
\footnote{For simplicity, the score for a battle between two models $A$ and
$B$ on instruction $x$ is written $s(x) = s(x, A) - s(x,B)$, where $s(x,A)$
denotes the judge's score for completion $A$ on $x$. 
}
We obtain this score by grading each completion on six criteria: adherence,
helpfulness, factuality, completeness, clarity, and fluency. Each criterion is
scored on a $1$--$10$ scale, and we average the six per-criterion differences
to obtain the battle-level score difference. To reduce position bias, we also evaluate
the swapped presentation order and average both model-oriented score differences, as done
for instance in \cite{li2024arenahard}.

The label used to fit BT in \cref{eq:bt-mle} is set to $1$, $0.5$, or $0$
when the final score difference favors $A$, is tied, or favors $B$, respectively.
When the score sign changes after swapping the two completions, we record a tie.
Appendix~\ref{app:judge-protocol} gives the full prompt, score extraction,
score parsing, balanced sampling protocol, and per-language corpus statistics.

\subsection{Split Conformal Prediction for Elo Estimation}
\label{sec:conformal-method}

Our goal is to estimate a new model's human Elo rating with an interval that is
both valid and useful. We use split conformal prediction~\citep{vovk2005algorithmic,
lei2018distribution} on LLM--human Elo residuals. The target of inference is each model's
Elo rating on the \emph{fitted human leaderboard} --- an observable quantity from Chatbot
Arena votes --- not an unobserved latent skill parameter. For each calibration model $i\in M$,
let $\eloHumani{i}$ denote its human Elo and
$\eloLLMi{i}$ its LLM-derived Elo, and define the per-model
\textbf{signed residual}
\begin{equation}
  \varepsilon_i \;=\; \eloLLMi{i} - \eloHumani{i}.
  \label{eq:elo-residual}
\end{equation}
Let $\widehat{SE}_i$ denote the bootstrap standard error of $\eloLLMi{i}$
across battle subsamples. Following the idea of normalized nonconformity
scores for regression~\citep{papadopoulos2008normalized}, we scale each
absolute Elo residual by its bootstrap standard error:
\begin{equation}
  S_i = |\varepsilon_i| / \widehat{SE}_i,
  \label{eq:conformal-score}
\end{equation}
so the bootstrap standard error sets the local scale of the interval while conformal
calibration supplies the coverage correction. The bootstrap is taken \emph{within}
each model's own battles, so $\widehat{SE}_i$ captures battle-level sampling noise but
not any systematic per-model bias of the Elo ruler --- structural distortion is preserved
across resamples rather than averaged out.

At miscoverage level $\alpha$, the conformal interval for a new model with
LLM-derived Elo $\eloLLMi{n+1}$ and bootstrap standard error
$\widehat{SE}_{n+1}$ is
\begin{equation}
  C \;=\;
  \bigl[
    \eloLLMi{n+1} - \hat{q}\,\widehat{SE}_{n+1},
    \;
    \eloLLMi{n+1} + \hat{q}\,\widehat{SE}_{n+1}
  \bigr],
  \label{eq:conformal-interval}
\end{equation}
where $n$ is the calibration-pool size and $\hat q$ is the
$\lceil(1{-}\alpha)(n{+}1)\rceil$-th smallest value of the calibration
nonconformity scores $\{S_i\}_{i=1}^n$. Under exchangeability of calibration
and test scores, this interval achieves marginal coverage of at least
$1{-}\alpha$. Whether judge-derived Hard-Elo
residuals satisfy that condition --- and whether the resulting intervals are
tight enough to be useful --- is the question we take up in
Section~\ref{sec:anatomy}.

\section{The Cost of Ignoring the Judge Uncertainty}
\label{sec:anatomy}

The practical question is whether standard ternary-label Elo is reliable enough
for estimating human Elo directly. It is not: Hard-Elo often recovers the ranking, but it
mis-scales the Elo axis and leaves structured residuals that make conformal
intervals unnecessarily wide.

\subsection{Failure Signature: Good Rankings, Bad Ruler, Wide Intervals}
\label{sec:baseline-residuals}

Table~\ref{tab:judges} separates three ways a judge can agree with humans:
battle-level agreement, leaderboard rank fidelity, and Elo scale fidelity.
This separation is the first failure signature of Hard-Elo. The judges often
recover the ordering of models, but their Elo distances remain substantially
mis-scaled.
We measure battle-level agreement with Cohen's $\kappa$, rank fidelity with
Spearman $\rho$ against the human Elo leaderboard, and scale fidelity with
held-out Elo MAE. We also report the mean absolute score difference $|s|$ as a
simple measure of judge signal strength. Raw agreement and free-marginal
$\kappa$ show the same distinction; when both sides select a winner,
agreement is $71.9$--$77.5\%$ (Appendix~\ref{app:agreement-reference}).

\begin{table}[t]
\centering
\small
\setlength{\tabcolsep}{4pt}
\caption{\textbf{Hard-Elo diagnostics across judges.}
Mean absolute score difference $|s|$ measures signal strength;
$\kappa$ measures battle-level agreement; Spearman $\rho$ measures rank
fidelity; and held-out Elo MAE measures scale fidelity.
\textbf{Bold}: best value per column.
Raw three-way agreement is $43.8$--$48.5\%$ and Randolph's $\kappa$ is
$0.156$--$0.227$; per-judge diagnostics appear in
Appendix~\ref{app:agreement-reference}.}
\label{tab:judges}
\begin{tabular}{lrrrr}
\toprule
& Signal & Pairwise & Leaderboard & Scale \\
\cmidrule(lr){2-2}\cmidrule(lr){3-3}\cmidrule(lr){4-4}\cmidrule(lr){5-5}
Judge & Mean $|s|$ & $\kappa\uparrow$ & $\rho\uparrow$ & MAE $\downarrow$ \\
\midrule
DeepSeek-V3.2  & \textbf{2.08} & 0.182 & 0.943 & 63.4 \\
Qwen3.5-27B    & 1.62 & \textbf{0.231} & 0.967 & 46.0 \\
Gemma4-E4B     & 1.56 & 0.169 & 0.926 & 48.2 \\
Gemma4-26B-A4B & 1.55 & 0.230 & \textbf{0.976} & 55.9 \\
Qwen3-32B      & 1.51 & 0.162 & 0.968 & \textbf{27.5} \\
Llama-3.3-70B  & 1.49 & 0.147 & 0.868 & 43.9 \\
GPT-OSS-20B    & 1.40 & 0.185 & 0.956 & 34.5 \\
GPT-OSS-120B   & 1.34 & 0.200 & 0.968 & 47.4 \\
\bottomrule
\end{tabular}
\end{table}

The table shows that rank fidelity and scale fidelity come apart. Even the
lowest-$\kappa$ judge, Llama-3.3-70B, still recovers much of the ordering
($\rho = 0.868$). But all judges incur substantial Elo MAE, from $27.5$ to
$63.4$ Elo. Thus the problem is not simply that judges cannot rank models; it
is that the ruler induced by Hard-Elo has the wrong scale. This has an immediate
operational consequence: BT's win-probability $\sigma(\theta_i - \theta_j)$ is
systematically overconfident, since judge-side $|\theta_i - \theta_j|$ exceeds its
human counterpart and $\sigma(\cdot)$ saturates closer to $0$ or $1$ than the
underlying preference data warrant. The same distortion propagates downstream:
split-conformal coverage remains broadly close to the $90\%$ nominal target,
but the resulting median intervals span $132$--$261$ Elo under Hard-Elo
(Table~\ref{tab:conformal-widths}; Figure~\ref{fig:teaser}c), too wide for
operational model comparison. Coverage is not the bottleneck; the ruler is.
The question is where this distortion comes from.

\subsection{Root Cause: Collapsing Uncertainty with Point Labels}
\label{sec:why-structured}

When fitting the BT model, the common practice is to pass discrete labels encoding whether the battles resulted in a win, a tie or a loss. This is natural when using human annotations since human labels are often provided in this way to simplify human labour\footnote{Human uncertainty can be estimated with repeated or richer annotations, but this substantially increases annotation cost; for instance reassessing hard labels for ImageNet required a dedicated re-annotation effort~\citep{beyer2020imagenet}.}. However, LLM judges often output scores whose difference indicates not only which completion was preferred, but also how strong that preference was. This score-difference information is collapsed to a deterministic point label. This therefore treats identically battles with large score differences (e.g.\ a battle won with a score of 8 against 2) compared to battles with small score differences (e.g.\ a battle won with a score of 9 against 8).

This information loss matters because Hard-Elo cannot distinguish close
contests from decisive ones. Using human Elo quartiles as a reference, we find
that battles between similarly strong models have much smaller judge score differences
than battles across large strength differences: across judges, top-quartile
vs.\ top-quartile battles have median $|s| \approx 0.61$, compared with
$\approx 1.54$ for top-quartile vs.\ bottom-quartile battles. Hard-Elo treats
both cases as the same unit win or loss. One could try to address narrow
score differences by thresholding them into ties, but this also loses information: a
small, stable preference between two strong models is still useful for ranking
them. The problem is not that narrow wins should be ignored, but that Hard-Elo
gives them the same target as decisive wins. Known battle-level biases, such as
position, verbosity, and self-preference, are measurable in our data, but they
do not by themselves explain the monotone residual ramp
(Appendix~\ref{app:bias-details}). The most consistent explanation is therefore
informational: Hard-Elo discards the local uncertainty encoded in the score
gap.

\subsection{Structured Residuals Inflate the Conformal Quantile}
\label{sec:residual-structure-hard}

A bootstrap standard error captures finite-sample variation from resampling a
model's battles, but it does not remove systematic Elo-scale bias that repeats
across those resamples.

\begin{figure}[t]
\centering
\includegraphics[width=\linewidth]{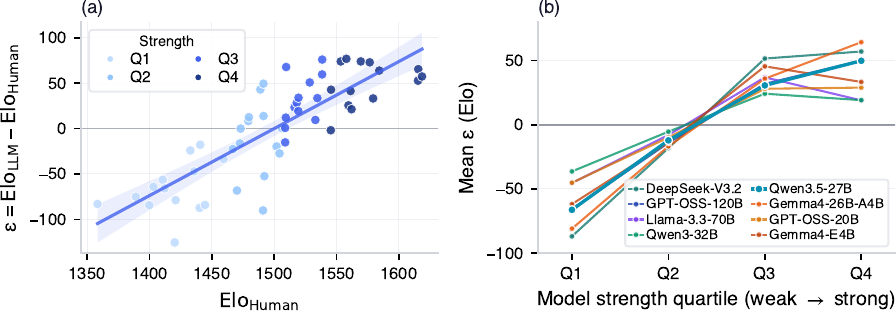}
\caption{\textbf{Hard-Elo residuals are strength-correlated on every judge.}
(a)~Signed residual $\varepsilon$ vs.\ human Elo for Qwen3.5-27B; points coloured by
strength quartile (light = weak; dark = strong), with linear fit and $95\%$ band.
(b)~Mean residual by $\eloHuman$ quartile, one line per judge.}
\label{fig:residual-heteroscedastic}
\end{figure}

Figure~\ref{fig:residual-heteroscedastic} shows that Hard-Elo residuals are
not exchangeable across model strength. Across all judges, the signed residual
$\varepsilon_i = \eloLLMi{i} - \eloHumani{i}$ correlates positively with human
Elo, with per-judge Pearson correlations from $+0.49$ to $+0.90$. Panel~(a)
shows one representative judge, Qwen3.5-27B: weak models lie mostly below zero
and strong models mostly above zero. Panel~(b) aggregates the same pattern by
human-Elo quartile, where Q1 denotes the weakest quartile and Q4 the strongest:
every judge moves from negative mean residuals in the lower quartiles to
positive mean residuals in the upper quartiles.

This structure directly affects the conformal intervals. Because
$\widehat{SE}_i$ captures resampling noise but not systematic strength-dependent
bias, the normalized scores $S_i = |\varepsilon_i|/\widehat{SE}_i$ remain
heterogeneous across the strength axis. The conformal quantile $\hat q$ must
cover these large normalized residuals, and every interval inherits the same
$\hat q$. The lever for narrower intervals at the same nominal coverage
is therefore not the conformal algorithm; it is the residuals that feed it. This
points at the diagnosis underlying the section: the problem is not BT itself,
but the hard targets it is fit to.

\section{From Local to Global Uncertainty for Calibrated Elo Estimation}

\label{sec:softelo}

\subsection{Calibrating the Score-to-Probability Map}
\label{sec:soft-method}

We now test the minimal fix motivated by the diagnosis above: keep the BT
model fixed, but stop collapsing the judge's score difference into a discrete
battle target. Concretely, Soft-Elo replaces the hard label $y_{ij}(x)$ in the
BT log-likelihood (Eq.~\ref{eq:bt-mle}) with the \textbf{soft target}
\begin{equation}
  \tilde y(x) \;=\; P[B \prec A \mid x]
  \;=\; \sigma\!\bigl(\beta\,s(x)\bigr) \;\in\; (0,\,1),
  \label{eq:soft-target}
\end{equation}
where $\beta > 0$ is a temperature that maps the score to a preference
probability: small $\beta$ keeps every target near $\tfrac{1}{2}$; large
$\beta$ concentrates the fit on battles with large score differences and recovers the behavior of hard labels.

We fit this temperature inside the same leave-one-model-out evaluation used for
Elo estimation. When model $m$ is held out, $\mathcal{B}^*_{-m}$ contains only
human non-tie battles between models other than $m$, so no human label
involving $m$ is used to calibrate the score map. The per-fold MLE
\begin{equation}
  \beta_m^*
  \;=\;
  \arg\max_{\beta}
  \sum_{(i,j,x)\,\in\,\mathcal{B}^*_{-m}}
  \Bigl[
    y^*_{ij}(x)\log\sigma\!\bigl(\beta\,s(x)\bigr)
    + \bigl(1{-}y^*_{ij}(x)\bigr)\log\bigl(1{-}\sigma\!\bigl(\beta\,s(x)\bigr)\bigr)
  \Bigr]
  \label{eq:temp-mle}
\end{equation}
varies negligibly across folds (per-judge std $\le 0.005$;
Appendix~\ref{app:beta-stability}); we report the cross-fold mean
$\bstar = \frac{1}{M}\sum_m \beta_m^*$ in Table~\ref{tab:soft-results}.
Human ties are excluded because Eq.~\ref{eq:temp-mle} calibrates a binary
preference map. After fitting $\beta_m^*$, the soft target is applied to all
judge-scored battles for the held-out model, including near-ties.

\begin{figure}[t]
\centering
\includegraphics[width=\linewidth]{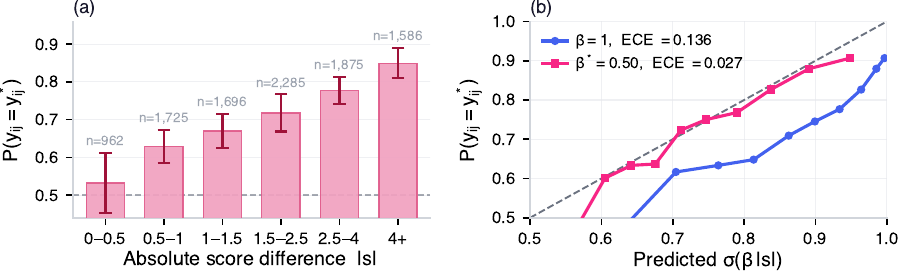}
\caption{\textbf{Score differences become calibrated preference probabilities after fitting $\beta$.}
(a)~Cross-judge agreement rate $P[y_{ij} = y^*_{ij}]$ on non-tied battles,
binned by $|s|$. Bars: cross-judge mean; error bars: $\pm 1\,\mathrm{SD}$;
$n$: number of non-tied battles per judge per bin.
(b)~Reliability diagram for $\sigma(\beta\,|s|)$ on Qwen3-32B (lowest-ECE judge);
diagonal = perfect calibration.}
\label{fig:margin-agreement}
\end{figure}

Figure~\ref{fig:margin-agreement} validates the score difference on non-tied
comparisons, where both the human and judge choose a side rather than a tie.
On this subset, judge--human agreement rises monotonically with $|s|$, from
near-chance ($53\%$) at small score differences to $85\%$ at large score differences ($|s|>4$),
showing that score differences carry local uncertainty information. BT, however,
needs this signal on a probability scale: its likelihood treats each battle
target as the probability that $A$ beats $B$ on prompt $x$. The temperature
$\beta$ provides this calibration step, turning the raw score difference into
the soft target
$\tilde y(x)=\sigma(\beta s(x))$.
Figure~\ref{fig:margin-agreement}b illustrates this on a representative
judge (Qwen3-32B): the probability obtained with $\sigma(s)$ ($\beta = 1$) is over-confident (Expected Calibration Error, ECE $=0.14$), while the probability obtained after fitting $\sigma(\bstar s)$ has much better calibrated uncertainty (ECE $=0.03$). 

We then fit the same BT objective as in Hard-Elo (Eq.~\ref{eq:bt-mle}),
replacing each hard target $y_{ij}(x)$ with the calibrated soft target from
Eq.~\ref{eq:soft-target}, using the fold-specific $\beta_m^*$ for each
held-out model. We call this pipeline
\textbf{Soft-Elo}: the Bradley--Terry model and Elo conversion are unchanged,
but the battle targets are softened using the calibrated judge score difference.
The held-out Elo-estimation protocol and split conformal construction are unchanged. This
is closely related in spirit to reward-modeling work that replaces binary
preference labels with ties or ordinal feedback when richer supervision is
available~\citep{rewardties2024, liu2024ordinal, afsharrad2026ordinal}.

\subsection{Held-Out Elo Estimation}
\label{sec:soft-results}

\begin{figure}[t]
\centering
\includegraphics[width=\linewidth]{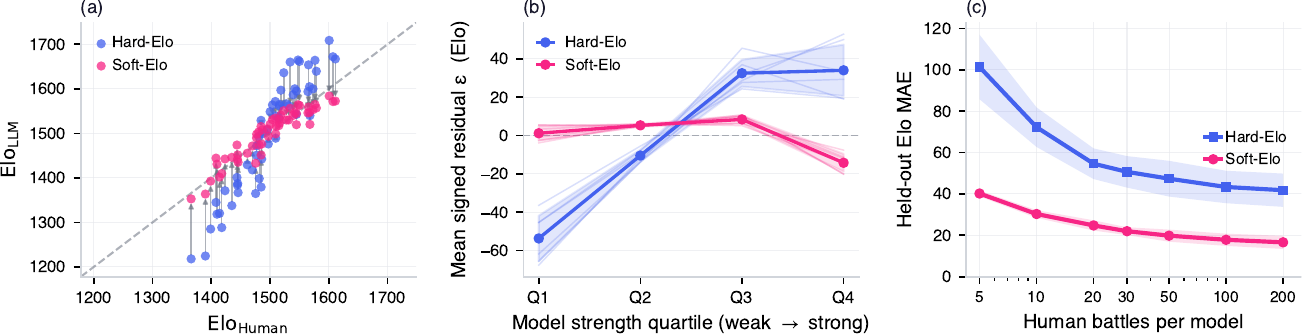}
\caption{\textbf{Soft-Elo corrects the stretched ruler and is more sample-efficient than Hard-Elo.}
(a)~Held-out judge Elo vs.\ human Elo for DeepSeek-V3.2 (largest Hard-Elo
MAE); Hard (blue) and Soft (pink) joined by arrows; dashed line: identity.
(b)~Mean signed residual $\varepsilon$ by Elo quartile.
(c)~Cross-judge mean held-out Elo MAE vs.\ annotation budget: Soft-Elo
beats Hard-Elo at every budget, with the largest gap in the small-data
regime.
Shaded bands in (b) and (c): $\pm 1\,\mathrm{SD}$ across the eight judges.
\label{fig:softelo-calibration}
}
\end{figure}

Having converted score differences into calibrated battle targets, we next ask
whether this local calibration improves the Elo estimates themselves. Using the
same leave-one-model-out protocol as Hard-Elo, we estimate each held-out model's
Elo from LLM-scored battles against anchor models and compare it to the human
Elo reference. Table~\ref{tab:soft-results} reports the resulting scale-fidelity
gains, and Figure~\ref{fig:softelo-calibration} shows how the residual structure
changes.

Soft-Elo substantially improves scale fidelity across all eight judges.
Held-out Elo MAE falls by $39$--$73\%$, with mean MAE dropping from $45.9$
to $\MAEEloAvg{}$~Elo. Ranking remains stable: the change in Spearman rank
correlation $\Delta\rho = \rho_\mathrm{soft} - \rho_\mathrm{hard}$ stays
between $-0.011$ and $+0.012$ across judges (Table~\ref{tab:soft-results}). Soft-Elo therefore corrects
the Elo scale without materially changing the leaderboard order.

\begin{table}[t]
\centering
\small
\caption{Hard-Elo vs.\ Soft-Elo results for held-out Elo estimation across
the eight judges, sorted by MAE reduction.
$\bstar$ is the cross-fold mean defined in Eq.~\ref{eq:temp-mle}.
$\Delta\rho$ reports the change in Spearman rank correlation relative to Hard-Elo.}
\label{tab:soft-results}
\setlength{\tabcolsep}{4pt}
\begin{tabular}{lrrrrr}
\toprule
& & \multicolumn{2}{c}{Elo MAE $\downarrow$} & & \\
\cmidrule(lr){3-4}
Judge & $\bstar$ & Hard & Soft & $\Delta$MAE\% & $\Delta\rho$ \\
\midrule
DeepSeek-V3.2  & 0.36 & 63.4 & 17.1 & $\mathbf{73\%}$ & $+0.003$ \\
Gemma4-26B-A4B & 0.54 & 55.9 & 15.7 & $72\%$ & $-0.011$ \\
Qwen3.5-27B    & 0.60 & 46.0 & 13.6 & $70\%$ & $+0.003$ \\
GPT-OSS-120B   & 0.58 & 47.4 & 14.4 & $70\%$ & $-0.002$ \\
Gemma4-E4B     & 0.38 & 48.2 & 21.0 & $57\%$ & $+0.012$ \\
Llama-3.3-70B  & 0.41 & 43.9 & 24.5 & $44\%$ & $+0.004$ \\
GPT-OSS-20B    & 0.46 & 34.5 & 19.8 & $42\%$ & $+0.009$ \\
Qwen3-32B      & 0.50 & 27.5 & 16.7 & $39\%$ & $+0.005$ \\
\bottomrule
\end{tabular}
\end{table}

Figure~\ref{fig:softelo-calibration} makes the correction visible. In
panel~(a), Soft-Elo moves held-out model estimates toward the identity line
between judge Elo and human Elo. In panel~(b), the strength-correlated residual
ramp from Section~\ref{sec:residual-structure-hard} is flattened, especially at
the weak and frontier quartiles.

Appendix~\ref{app:extra-datasets} reports results on LMArena 140K and ComparIA,
including the Qwen failure cases under the corpus shift.
Appendices~\ref{app:annotation-budget} and~\ref{app:calibration-sensitivity}
examine annotation budgets, prompt sensitivity, and human-reference uncertainty.

These smaller and less structured residuals also determine the uncertainty
intervals around new-model Elo estimates, which we evaluate next.

\subsection{Model-Level Conformal Intervals}
\label{sec:conformal}
\label{sec:cp-application}
\label{sec:cp-coverage}

Soft-Elo improves the Elo estimate, but the LLM-derived Elo remains a proxy for
the human Elo. We therefore wrap each new-model Elo estimate with a model-level
conformal interval. Using the split conformal construction from
Section~\ref{sec:conformal-method}, we calibrate LLM--human Elo residuals across
held-out models and return an interval on the human Elo axis with
distribution-free marginal coverage under exchangeability.

We evaluate this construction retrospectively via leave-one-model-out over the
LMArena pool. In a forward deployment, the calibration pool contains models with
both human and LLM-derived Elo estimates; a new model requires only LLM-scored
battles, and receives a conformal interval for its human Elo rating. For each
judge, we run five random model-level splits of the $55$ held-out models, using
$27$ models for conformal calibration and $28$ for testing. The per-model scale
$\widehat{SE}_i$ is estimated with $20$ within-model battle bootstrap resamples.

Soft-Elo's smaller residuals propagate into the conformal half-width
$\hat q \cdot \widehat{SE}_i$ (Eq.~\ref{eq:conformal-interval}), either through
the conformal quantile $\hat q$ or the per-model bootstrap scale
$\widehat{SE}_i$; Appendix~\ref{sec:cp-decomposition} decomposes these effects.
Table~\ref{tab:conformal-widths} shows the result: Soft-Elo reduces median
split-conformal width by $39$--$70\%$ while keeping empirical coverage near the
$90\%$ nominal target ($88.6$--$97.9\%$ for Hard-Elo; $92.1$--$96.4\%$ for
Soft-Elo). Figure~\ref{fig:teaser}c illustrates the same model-level contraction
for a representative judge.

\begin{table}[t]
\centering
\small
\caption{Split-conformal interval widths and empirical coverage at $90\%$ nominal
coverage across the eight judges.
Each entry: mean median width / mean coverage across 5 random calibration/test
splits; bracketed: 2.5th--97.5th percentiles across the splits.
\textbf{Bold}: narrowest soft-label width.}
\label{tab:conformal-widths}
\setlength{\tabcolsep}{4pt}
\begin{tabular}{lrrrrrr}
\toprule
& \multicolumn{3}{c}{Hard-Elo} & \multicolumn{3}{c}{Soft-Elo} \\
\cmidrule(lr){2-4}\cmidrule(lr){5-7}
Judge & Cov.\ & Mean median & Range & Cov.\ & Mean median & Range \\
\midrule
DeepSeek-V3.2  & 95.0\% & 261.0 & [239, 285] & 92.9\% & \phantom{0}78.0 & [58, 89]  \\
GPT-OSS-120B   & 97.9\% & 243.9 & [212, 266] & 92.9\% & \phantom{0}75.1 & [60, 99]  \\
Qwen3.5-27B    & 90.0\% & 212.2 & [143, 257] & 95.7\% & \phantom{0}\textbf{74.4} & [69, 85]  \\
Gemma4-26B-A4B & 95.7\% & 219.6 & [198, 236] & 95.7\% & \phantom{0}78.2 & [66, 86]  \\
Llama-3.3-70B  & 94.3\% & 249.2 & [202, 366] & 94.3\% & 132.6 & [116, 158]  \\
GPT-OSS-20B    & 95.7\% & 186.8 & [165, 225] & 94.3\% & 102.9 & [86, 118]  \\
Qwen3-32B      & 88.6\% & 131.9 & [79, 184] & 92.1\% & \phantom{0}78.0 & [61, 102]  \\
Gemma4-E4B     & 90.0\% & 235.2 & [136, 285] & 96.4\% & 143.1 & [104, 171]  \\
\bottomrule
\end{tabular}
\end{table}

Appendix~\ref{app:deployment-shift} examines calibration transfer across corpora
and Appendix~\ref{app:temporal-arrival} evaluates models arriving after a
chronological cutoff.

\section{Conclusion and Limitations}
\label{sec:discussion}

We introduced Soft-Elo for estimating human Elo ratings from LLM-judged
battles. At the benchmarking level, we estimate Elo directly from battles
against multiple opponents rather than reporting win rate against a fixed
baseline. At the local level, Soft-Elo replaces discrete win/loss/tie targets
with calibrated preference probabilities derived from score differences,
improving scale fidelity while preserving rank order. At the global level, we
apply split conformal prediction to LLM--human Elo residuals, yielding conformal
intervals for new models' human Elo ratings.

The practical payoff is that uncertainty is calibrated at the level developers
actually use: model-level Elo estimates, not individual instructions. Soft-Elo
reduces the residuals that feed the conformal construction, so the resulting
intervals are substantially narrower at the same nominal coverage. The method
therefore keeps the standard Bradley--Terry and split conformal machinery, but
changes the information passed into it. The interval-width decomposition and coverage limitations are discussed in
Appendix~\ref{app:interval-analysis}.

Two limitations remain. First, Soft-Elo uses only the uncertainty encoded in the
judge's score difference; it does not model epistemic uncertainty, hallucinated
scores, prompt ambiguity, or judge instability, or remove same-family preference bias. Second, the conformal guarantee
is marginal and relies on exchangeability between calibration and future models.
Prompt-distribution, model-family, or temporal shifts can violate this
condition, and marginal coverage does not imply coverage within every
subgroup. The tested shifts and deployment diagnostics are reported in
Appendix~\ref{app:deployment-shift}; shifted deployments should re-estimate the score-difference
diagnostic and the conformal calibration set before trusting the intervals.

\paragraph{Broader Impact Statement.}
Widespread reliance on LLM judges without accounting for their biases or
uncertainty can distort model rankings and mislead the developers and users
relying on them. We believe the proposed approach is a net positive: by
surfacing calibrated uncertainty at the model level --- the granularity at
which practitioners make deployment decisions --- it encourages better-informed
leaderboard interpretations.

\section*{Acknowledgements and Disclosures}
We thank the anonymous reviewers for their constructive feedback, which helped improve the paper's clarity and empirical evaluation.

This research is cofunded by the European Union (GA No 101195233). Views and opinions expressed are however those of the OpenEuroLLM consortium only and do not necessarily reflect those of the European Union or Digital Europe Programme. Neither the European Union nor the granting authority can be held responsible for them.

\paragraph{Competing interests and additional support.}
The authors declare no competing interests and received no additional
funding or third-party support beyond the funding acknowledged above.

\bibliographystyle{unsrt}
\bibliography{references}

\clearpage
\appendix

\section{Evaluation Protocol and Resources}
\label{app:judge-protocol}

\subsection{Battle Corpora and Per-Language Coverage}
\label{app:battle-corpora}

The main experiments use pairwise battles from \textbf{LMArena 100K}, a
multilingual subset of Chatbot Arena~\citep{chiang2024chatbot}. This corpus
defines the primary model pool, judge comparisons, Soft-Elo calibration, and
conformal interval results reported in the main paper.

We use two additional corpora only for replication and stress testing.
\textbf{LMArena 140K} extends the Arena pool with more recent model releases
under similar prompt characteristics (Sec.~\ref{app:lmsys140k}).
\textbf{ComparIA} is a French-language pairwise corpus and tests whether the
score-difference signal transfers under a stronger prompt-distribution shift
(Sec.~\ref{app:comparia}).
The LMArena corpora are distributed through Hugging Face\footnote{\href{https://huggingface.co/datasets/lmarena-ai/arena-human-preference-100k}{LMArena 100K};
\href{https://huggingface.co/datasets/lmarena-ai/arena-human-preference-140k}{LMArena 140K}.},
and ComparIA votes are also publicly available on Hugging Face.\footnote{\href{https://huggingface.co/datasets/ministere-culture/comparia-votes}{ComparIA votes}.}
Licensing and redistribution details for these corpora and our derived
annotation files are given in Appendix~\ref{app:licenses}.

Figure~\ref{fig:per-language-counts} reports per-language instruction counts
across the three corpora. LMArena 100K and 140K are English-dominated
multilingual mixtures (${\approx}54$--$58\%$ English, with substantial Russian,
Chinese, German, and Polish tails); ComparIA is exclusively French.

\subsection{Balanced Per-Model Sampling}
We annotate approximately $25{,}000$ pairwise battle rows per judge on
LMArena 100K. For the LMArena 140K and ComparIA replication corpora, we use
the same balanced sampling rule with comparable per-judge row budgets. For the
main LMArena 100K pool, because each battle contains two models, this
corresponds to approximately $2|\mathcal{B}|/M$ model appearances on average
across the $M = 55$ models. In the paper we report sampling budgets using the
normalized per-model battle-row budget
\[
  b = |\mathcal{B}|/M,
\]
so the full $25{,}000$-battle annotation set corresponds to
$b \approx 455$ judged battle rows per model, or about $910$ model
appearances per model on average.

Sampling is balanced with respect to model coverage as far as the available
Arena comparison graph allows: we avoid concentrating the annotation budget on
a small subset of frequently appearing models, but the realized number of
appearances is not identical for every model because the underlying pairwise
pool is itself uneven. Robustness curves over $b$ should therefore be read as
varying the judge-query budget normalized by the number of models, rather than
as enforcing exactly $b$ appearances for every model.

\begin{figure}[h]
  \centering
  \includegraphics[width=0.92\linewidth]{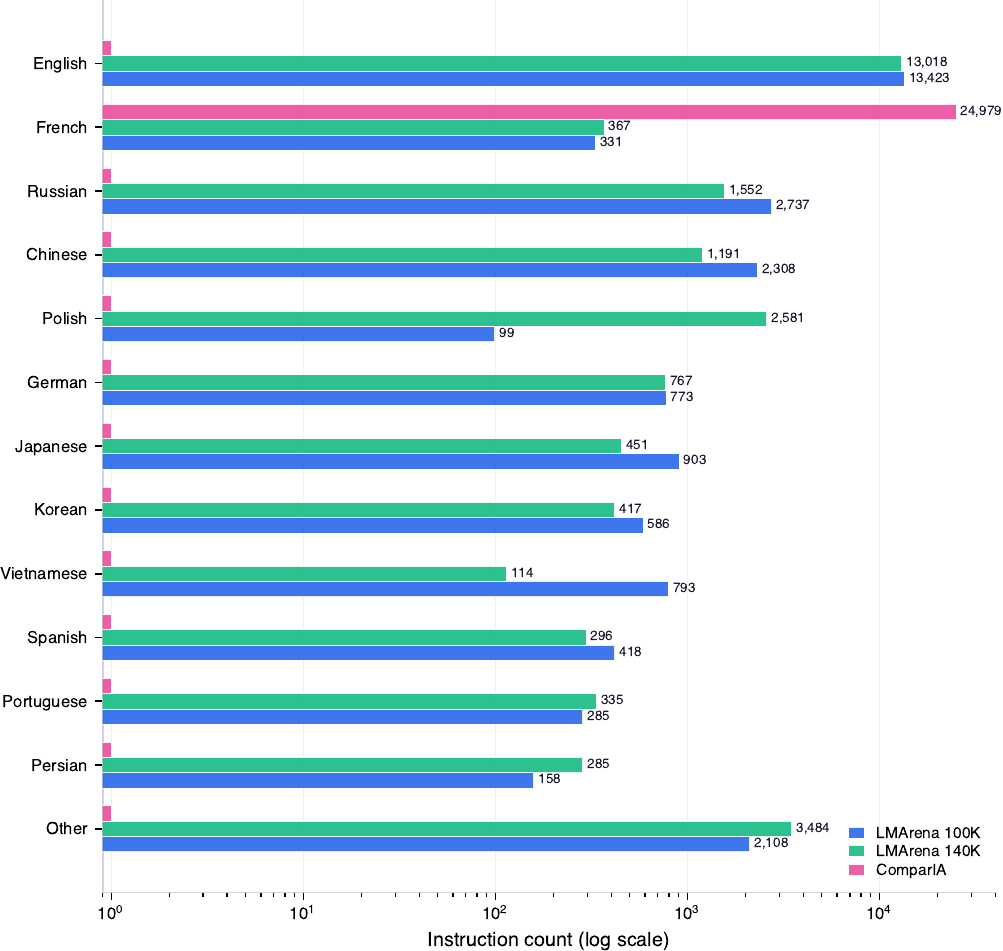}
  \caption{\textbf{LMArena spans ${\sim}80$ languages with English dominant;
    ComparIA is exclusively French.}
    Per-language instruction counts across the three pairwise battle
    corpora, on a logarithmic horizontal axis. LMArena 100K (blue) and
    LMArena 140K (green) cover ${\sim}80$ languages; ComparIA (pink) is
    fully French. The corpora are introduced in Section~\ref{sec:judges};
    140K and ComparIA are used for the cross-corpus replication
    in Sections~\ref{app:lmsys140k} and~\ref{app:comparia}.}
  \label{fig:per-language-counts}
\end{figure}

\paragraph{Shared battles.}
\label{app:shared-battles}
The committee and human-reference bootstrap analyses use 18,158 battles
shared by all eight judges, covering the same model pool. This intersection
is smaller than the per-judge annotation sets described above.

\subsection{System Prompt}

Judges receive the following system prompt; the placeholders are filled
with the criterion descriptions below, an optional explanation
instruction, and a JSON schema example.

\begin{quote}\small\ttfamily
You are an expert evaluator comparing two AI assistant responses.\\
You will see an instruction and two completions (A and B).\\[4pt]
Score EACH completion on every criterion below, then give your overall preference.\\[4pt]
\{criteria\_block\}\\[4pt]
\{explanation\_block\}\\[4pt]
Return your output as JSON with the following structure:\\
\`{}\`{}\`{}json\\
\{example\_json\_pairwise\}\\
\`{}\`{}\`{}\\[4pt]
IMPORTANT:\\
- Score BOTH completions on ALL criteria.\\
- Do NOT let position (A vs B) bias your scores.\\
- Base your preference on the criteria scores and overall quality.\\
- Wrap your JSON in \`{}\`{}\`{}json \ldots{} \`{}\`{}\`{} tags.\\
- Your response MUST begin with \`{}\`{}\`{}json on the first line.\\
- Do NOT write any prose, analysis, or bullets before the JSON block.\\
- If you include any explanation, put it AFTER the JSON block, never before it.
\end{quote}

\subsection{Judge Generation Settings}

All judge calls use deterministic decoding with temperature $0.0$, top-$p=1.0$
(no nucleus filtering), and a maximum output budget of $4096$ tokens. The token
budget is used to allow the judge to complete the full JSON response; explanations,
when enabled, are placed after the JSON block and are not used by the scoring
pipeline.

\subsection{User Prompt Template}

\begin{quote}\small\ttfamily
\#\# Instruction\\
\{instruction\}\\[4pt]
\#\# Completion A\\
\{completion\_A\}\\[4pt]
\#\# Completion B\\
\{completion\_B\}\\[4pt]
Compare both completions. Score each on every criterion, then state your overall
preference. Start your response with the JSON scores immediately.
\end{quote}

\subsection{Scoring Criteria}

All criteria are scored on a $1$--$10$ scale.
Anchors at scores 10, 7, 4, and 1 are provided; intermediate values are interpolated
by the judge.

\begin{enumerate}[leftmargin=*, itemsep=8pt]

  \item \textbf{Adherence.}
        Follows the user's instructions and constraints precisely: required format,
        scope, style constraints, and any do/don't requirements. Penalize missing
        requested parts or deviating from constraints.\\[2pt]
        \begin{tabular}{@{}cl@{}}
          10 & Fully follows the user's instructions and constraints, including format and scope.\\
           7 & Mostly follows the request but misses some details or adds minor unnecessary content.\\
           4 & Follows only part of the request; misses important constraints or requested parts.\\
           1 & Does not follow the user's request in any meaningful way.
        \end{tabular}

  \item \textbf{Helpfulness.}
        Advances the user's goal with relevant, actionable content. Provides useful
        steps, options, or explanations tailored to the request. Penalize generic
        filler or non-responsive content.\\[2pt]
        \begin{tabular}{@{}cl@{}}
          10 & Directly solves the user's problem with highly useful, actionable, and relevant content.\\
           7 & Generally helpful and relevant, but misses some useful detail or optimization.\\
           4 & Partially helpful but vague, generic, or missing key actionable guidance.\\
           1 & Unhelpful or non-responsive to the user's goal.
        \end{tabular}

  \item \textbf{Factuality.}
        Information is correct and appropriately qualified. Avoids hallucinations and
        unwarranted specifics. If uncertain, expresses uncertainty and does not
        fabricate sources, citations, or details.\\[2pt]
        \begin{tabular}{@{}cl@{}}
          10 & Accurate and well-qualified throughout, with no fabricated or unsupported claims.\\
           7 & Mostly accurate, with only minor imprecision or insufficient qualification.\\
           4 & Contains multiple factual errors, overclaims, or likely hallucinated details.\\
           1 & Largely incorrect, fabricated, or misleading.
        \end{tabular}

  \item \textbf{Completeness.}
        Covers the key aspects of the request without major omissions. Addresses all
        sub-questions and important constraints. Penalize partial answers or skipped
        items.\\[2pt]
        \begin{tabular}{@{}cl@{}}
          10 & Covers all major parts of the request with no important omissions.\\
           7 & Covers the main request but misses some secondary details or sub-parts.\\
           4 & Only partially addresses the request; multiple important pieces are missing.\\
           1 & Fails to address the requested task in a complete or usable way.
        \end{tabular}

  \item \textbf{Clarity.}
        Well-organized, easy to follow, and unambiguous. Uses logical structure,
        headings/lists when helpful, and clear references. Penalize confusing
        organization or ambiguity.\\[2pt]
        \begin{tabular}{@{}cl@{}}
          10 & Clear, logically structured, and easy to follow with no ambiguity.\\
           7 & Mostly clear and organized, but has a few awkward transitions or ambiguities.\\
           4 & Noticeably hard to follow due to weak structure or unclear phrasing.\\
           1 & Confusing, disorganized, or difficult to understand.
        \end{tabular}

  \item \textbf{Fluency.}
        Language and presentation quality: fluent, readable, appropriately concise,
        and well-formatted. Tone is appropriate for the user/context.\\[2pt]
        \begin{tabular}{@{}cl@{}}
          10 & Fluent, natural, polished language with strong readability and appropriate tone.\\
           7 & Generally fluent and readable, with some awkward phrasing or minor disfluencies.\\
           4 & Frequent disfluencies or formatting issues that make reading noticeably harder.\\
           1 & Severely disfluent or poorly formatted to the point of harming comprehension.
        \end{tabular}

\end{enumerate}

\subsection{Score Extraction}

For a completion produced by model $A$ on instruction $x$, let $s_c(x,A)$
denote the judge score on criterion $c \in \mathcal{C}$, where
$\mathcal{C}$ contains the six criteria above. We first average across
criteria,
\[
  s(x,A) = \frac{1}{|\mathcal{C}|}\sum_{c \in \mathcal{C}} s_c(x,A).
\]
For a battle between models $A$ and $B$, we write the model-oriented score
difference compactly as $s(x) = s(x,A) - s(x,B)$, so positive values favor
$A$ and negative values favor $B$.

To reduce position bias, each battle is judged twice: once with $A$ shown
first and $B$ second, and once with the order swapped. We map both outputs
back to the same model-oriented score difference $s(x)$ and average the two values to
obtain the final scalar score used in the paper. The Hard-Elo label is then
$y_{ij}(x)=1$ if the final score difference favors $A$, $y_{ij}(x)=0$ if it favors
$B$, and $y_{ij}(x)=0.5$ for a tie. In addition, if the sign of the
model-oriented score difference changes between the original and swapped presentations,
we record the battle as a tie, since the judge's preference is not stable
under position swapping.

\subsection{Held-out evaluation algorithm}
\label{app:softelo-algorithm}

Algorithm~\ref{alg:softelo} gives the full pseudocode for the Soft-Elo
held-out evaluation summarised in Section~\ref{sec:softelo}. For brevity,
$\tilde y_{ij}(x) \equiv P[B \prec A \mid x] = \sigma(\bstar\,s(x))$
denotes the soft target (Eq.~\ref{eq:soft-target}); we use $\tilde y$ as
shorthand throughout this appendix.

\begin{algorithm}[h]
\caption{Soft-Elo held-out evaluation with per-fold temperature $\beta_m^*$ and two-step LOO Bradley--Terry fit.}
\label{alg:softelo}
\begin{algorithmic}[1]
\Require Battles $b \in \mathcal{B}$ with endpoints $i(b),j(b)$, prompt $x_b$, scores $s(x_b)$, and human labels $y^*_{i(b)j(b)}(x_b) \in \{0, \tfrac12, 1\}$; model set $\mathcal{M}$; L2 regulariser $\lambda$.
\For{each held-out model $m \in \mathcal{M}$}
   \State $\mathcal{B}_{-m} \gets \{b \in \mathcal{B} : i(b)\neq m,\ j(b)\neq m\}$, \; $\mathcal{T}_m \gets \{b \in \mathcal{B} : m \in \{i(b),j(b)\}\}$
   \State $\mathcal{B}^*_{-m} \gets \{b \in \mathcal{B}_{-m} : y^*_{i(b)j(b)}(x_b) \in \{0,1\}\}$ \Comment{anchor-only human non-ties}
   \State $\beta_m^* \gets$ MLE on $\mathcal{B}^*_{-m}$ via Eq.~\ref{eq:temp-mle}
   \State $\tilde y_{i(b)j(b)}(x_b) \gets \sigma\!\bigl(\beta_m^*\,s(x_b)\bigr)$ for all $b \in \mathcal{B}_{-m}\cup\mathcal{T}_m$
   \State $\hat{\boldsymbol\theta}^{\text{anch}} \gets$ BT-MLE on $\mathcal{B}_{-m}$ with target $\tilde y$ and regulariser $\lambda$ \Comment{anchor strengths}
   \State $\hat\theta_m \gets \displaystyle\arg\max_{\theta_m}$ BT log-likelihood on $\mathcal{T}_m$ with target $\tilde y$, holding anchors fixed at $\hat{\boldsymbol\theta}^{\text{anch}}$
   \State $\mathrm{err}_m \gets \bigl|\,\mathrm{Elo}(\hat\theta_m) - \eloHumani{m}\,\bigr|$ \Comment{$\eloHumani{m}$: same two LOO steps with target $y^*$ on $\mathcal{B}_{-m}, \mathcal{T}_m$}
\EndFor
\State \Return $\{\mathrm{err}_m\}_{m \in \mathcal{M}}$
\end{algorithmic}
\end{algorithm}

\subsection{Agreement metrics}
\label{app:agreement-reference}

Table~\ref{tab:agreement_diagnostics} extends Table~\ref{tab:judges} with raw
three-way agreement, Randolph's free-marginal $\kappa$, and decisive-battle
agreement. We use the same preference-to-class mapping as Table~\ref{tab:judges},
with a $\pm0.05$ tie band around $0.5$. Randolph's statistic replaces the
marginal-dependent chance term with $1/k$, giving
$\kappa_{\mathrm{free}}=(P_o-1/k)/(1-1/k)$ for observed agreement $P_o$
and $k=3$ classes. Its values remain close to Cohen's $\kappa$, so the low
three-way agreement is not explained by the chance-correction convention.
When both the human and judge select a winner, agreement rises to
$71.9$--$77.5\%$. Human tie rates are approximately $40\%$, whereas judge tie
rates range from $13\%$ to $47\%$. The tie boundary therefore matters when
interpreting battle-level agreement alongside aggregate leaderboard correlation.
\begin{table}[htbp]
\centering
\small
\setlength{\tabcolsep}{4pt}
\caption{Battle-level agreement on the main evaluation data. Decisive comparisons exclude ties from both human and judge labels.}
\label{tab:agreement_diagnostics}
\begin{tabular}{lrrrrr}
\toprule
Judge & Raw (\%) & Cohen $\kappa$ & Randolph $\kappa$ & Decisive (\%) & Decisive $\kappa$ \\
\midrule
Llama-3.3-70B & 43.8 & 0.147 & 0.156 & 73.0 & 0.459 \\
Qwen3-32B & 45.2 & 0.162 & 0.178 & 77.5 & 0.549 \\
Gemma4-E4B & 44.8 & 0.169 & 0.172 & 71.9 & 0.438 \\
DeepSeek-V3.2 & 45.2 & 0.182 & 0.178 & 74.2 & 0.484 \\
GPT-OSS-20B & 45.4 & 0.185 & 0.180 & 72.8 & 0.456 \\
GPT-OSS-120B & 46.2 & 0.200 & 0.192 & 74.1 & 0.482 \\
Gemma4-26B-A4B & 48.5 & 0.230 & 0.227 & 76.2 & 0.525 \\
Qwen3.5-27B & 47.3 & 0.231 & 0.209 & 74.5 & 0.491 \\
\bottomrule
\end{tabular}
\end{table}

\subsection{Compute Resources}
\label{app:compute-resources}

Open-weight judge annotations were generated on NVIDIA A100-SXM-64GB GPUs.
Jobs used between one and four GPUs depending on judge size: models above
roughly $30$B parameters used four GPUs, while smaller models used one or two
GPUs. DeepSeek-V3.2 annotations were generated through the OpenRouter API.
After annotation, the Bradley--Terry fits, temperature calibration, bootstrap
standard errors, split-conformal intervals, and plotting pipeline were run
locally as CPU-based post-processing over cached judge outputs on an
Apple-silicon laptop. Wall-clock times depend on GPU queueing and API latency
for annotation; the reported statistical analyses operate only on cached
annotation files.

\subsection{Licensing of Datasets and Models}
\label{app:licenses}

The pairwise human-preference corpora used in this work are publicly available:
LMArena 100K and 140K release user prompts under CC-BY-4.0, with model outputs
governed by each provider's own terms, and ComparIA votes are released under the
Etalab 2.0 Open License. Our released annotation files contain metadata, model
identifiers, preference labels, and judge scores, but exclude raw prompts and
completions to remain consistent with these terms. Of the eight judge models,
six are released under Apache 2.0
(Qwen3-32B, Qwen3.5-27B, Gemma 4-26B-A4B-it, Gemma 4-E4B-it, GPT-OSS-120B,
GPT-OSS-20B), one is released under the MIT License (DeepSeek-V3.2), and
Llama-3.3-70B-Instruct is released under the Llama 3.3 Community License
Agreement, whose only commercial-use restriction (services with $>$700M
monthly active users requiring Meta authorisation) does not apply to
non-commercial research use. All judge weights were obtained through their
publicly hosted releases and used in accordance with the above terms.

\section{Cross-Corpus and Cross-Lingual Evaluation}
\label{app:extra-datasets}

We test whether Soft-Elo's gains persist across corpora and languages, and
whether calibration can be reused across corpora. Following
Appendix~\ref{app:judge-protocol}, we first refit the pipeline on LMArena 140K
and ComparIA~\citep{termignon2026comparia}, then test transfer of the
temperature and conformal quantile. We also evaluate language subsets of the
original LMArena pool.

\subsection{LMArena 140K replication}
\label{app:lmsys140k}

LMArena 140K has a newer model pool and retains the Arena evaluation format
(Table~\ref{tab:extra-datasets}).
Across the overlapping judges,\footnote{Qwen3-32B is excluded from the
LMArena 140K replication: its context length cannot accommodate the longer
prompts in the 140K subset, and rerunning it on truncated prompts would not
be a fair comparison to the other judges.} Soft-Elo reduces Elo MAE by
$45$--$79\%$ while leaving rank correlation essentially unchanged.

\begin{table}[!ht]
\centering
\scriptsize
\caption{\textbf{Cross-corpus replication.}
We report battle-level agreement ($\kappa$), the fitted score-to-probability
temperature ($\bstar$), held-out Elo MAE, and rank correlation. LMArena 140K is
a newer Arena-style model pool; ComparIA is a French-language stress test.
$\kappa$ is computed on the filtered judged battles used by the pipeline.}
\label{tab:extra-datasets}
\setlength{\tabcolsep}{3pt}
\begin{tabular}{llrrrrrrr}
\toprule
& & & & \multicolumn{2}{c}{Elo MAE $\downarrow$} &
\multicolumn{2}{c}{Spearman $\rho$ $\uparrow$} & \\
\cmidrule(lr){5-6}\cmidrule(lr){7-8}
Corpus & Judge & $\kappa$ & $\bstar$ & Hard & Soft & Hard & Soft & $\Delta$MAE\% \\
\midrule
\multicolumn{9}{l}{\textit{LMArena 140K --- newer Arena-style model pool}} \\
\addlinespace[1pt]
LMArena 140K & Gemma4-E4B     & 0.177 & 0.41 & 88.0 & 18.8 & 0.924 & 0.921 & 79\% \\
LMArena 140K & Gemma4-26B-A4B & 0.213 & 0.60 & 77.5 & 19.2 & 0.938 & 0.945 & 75\% \\
LMArena 140K & DeepSeek-V3.2  & 0.173 & 0.40 & 70.9 & 17.4 & 0.929 & 0.930 & 75\% \\
LMArena 140K & Qwen3.5-27B    & 0.187 & 0.51 & 69.3 & 20.0 & 0.938 & 0.923 & 71\% \\
LMArena 140K & GPT-OSS-20B    & 0.058 & 0.51 & 67.5 & 22.8 & 0.900 & 0.908 & 66\% \\
LMArena 140K & GPT-OSS-120B   & 0.156 & 0.56 & 53.7 & 24.0 & 0.900 & 0.901 & 55\% \\
LMArena 140K & Llama-3.3-70B  & 0.131 & 0.47 & 56.7 & 31.1 & 0.754 & 0.752 & 45\% \\
\midrule
\multicolumn{9}{l}{\textit{ComparIA --- French-language stress test}} \\
\addlinespace[1pt]
ComparIA & Gemma4-26B-A4B & 0.156 & 0.34 & 161.3 & 30.6 & 0.770 & 0.785 & 81\% \\
ComparIA & Gemma4-E4B     & 0.134 & 0.30 & 146.2 & 32.3 & 0.717 & 0.746 & 78\% \\
ComparIA & GPT-OSS-120B   & 0.132 & 0.37 & 102.0 & 30.9 & 0.740 & 0.769 & 70\% \\
ComparIA & GPT-OSS-20B    & 0.132 & 0.34 & 107.9 & 33.0 & 0.719 & 0.748 & 69\% \\
ComparIA & Llama-3.3-70B  & 0.127 & 0.32 & \phantom{0}68.0 & 35.3 & 0.751 & 0.767 & 48\% \\
ComparIA & Qwen3.5-27B    & 0.102 & 0.14 & \phantom{0}82.9 & 47.8 & 0.845 & 0.660 & 42\% \\
ComparIA & Qwen3-32B      & 0.124 & 0.13 & \phantom{0}62.4 & 49.7 & 0.808 & 0.577 & 20\% \\
\bottomrule
\end{tabular}
\end{table}

\subsection{ComparIA: a score-signal stress test}
\label{app:comparia}

On the French-language ComparIA corpus, Soft-Elo reduces Elo MAE for all
overlapping judges in
Table~\ref{tab:extra-datasets}, but rank fidelity is no longer uniform. The
Gemma, GPT-OSS, and Llama judges keep stable or slightly improved $\rho$,
whereas the two Qwen judges lose substantial rank correlation.

\begin{figure}[t]
\centering
\includegraphics[width=\linewidth]{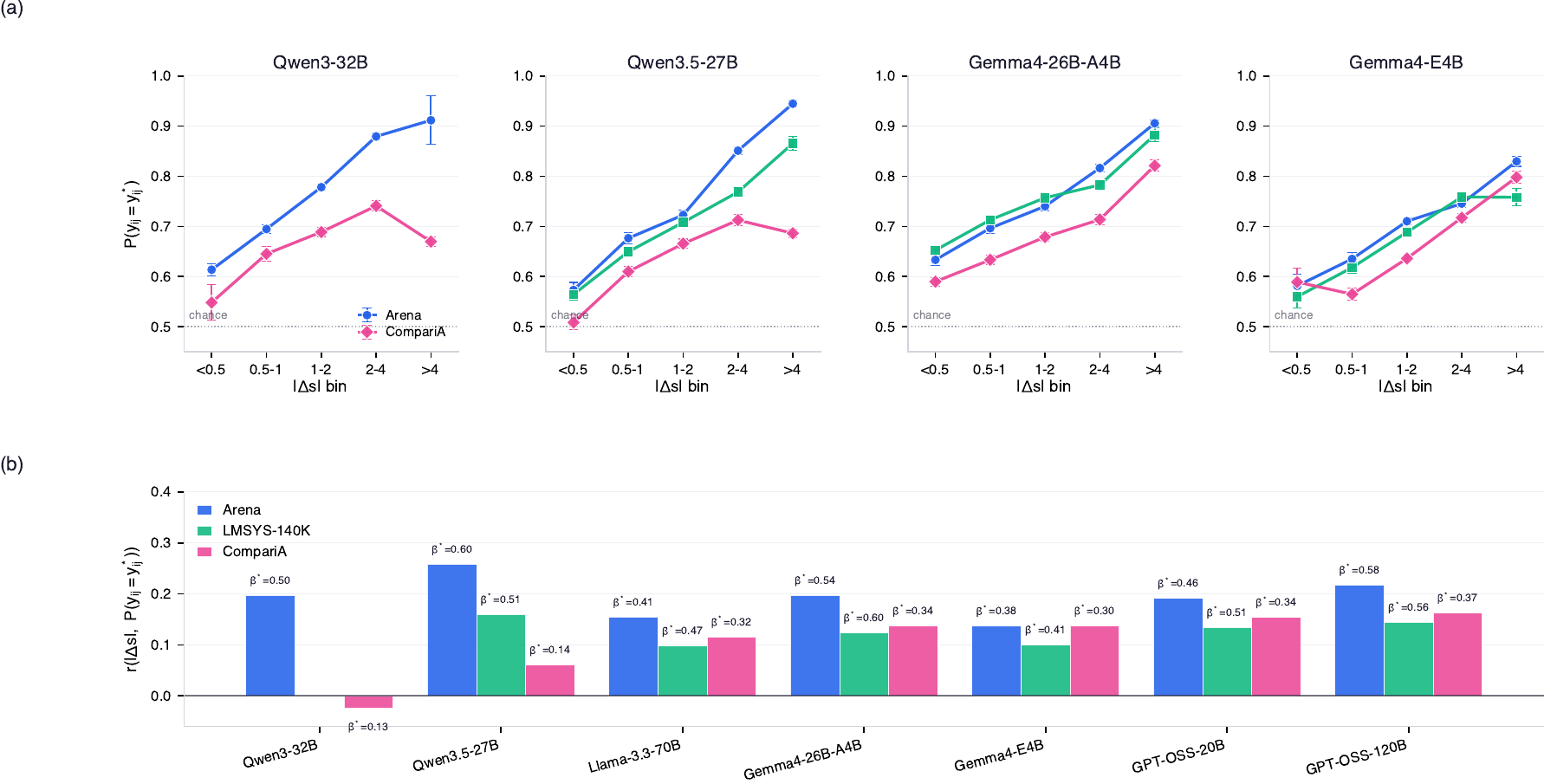}
\caption{\textbf{Soft-Elo helps when the score difference predicts agreement and
weakens when it does not.}
(a)~Per-judge agreement rate $P(y_{ij} = y^*_{ij})$ on decisive battles, binned
by score difference $|s|$, overlaid by corpus. Error bars are
$\pm 1$ binomial SE; bins with fewer than $30$ battles are omitted. LMArena 140K
runs cover only the two Gemma judges in this overlap.
(b)~Pearson $r\bigl(|s|, P(y_{ij} = y^*_{ij})\bigr)$ per
judge $\times$ corpus, annotated with the calibrated $\bstar$ for that
combination. Low correlation co-occurs with small $\bstar$.}
\label{fig:margin-corpora}
\end{figure}

For the two Qwen judges, $\bstar$ falls to $0.13$--$0.14$ and the
score-gap--agreement curve becomes nearly flat
(Figure~\ref{fig:margin-corpora}). Their soft targets compress the Elo scale,
lowering MAE but also reducing rank correlation. The per-judge calibration
errors show the same difficulty (Table~\ref{tab:calibration-corpora}).

Two checks help diagnose this failure. Qwen3-32B reverses its score-based
winner on $46.8\%$ of comparisons when presentation order is swapped and has
near-zero gap--agreement correlation. Qwen3.5-27B is more stable to order,
but its gaps remain weakly informative. Order reversal can be measured from
judge outputs alone; checking gap--agreement requires human-labeled calibration
battles. If either check indicates an unreliable signal, we recommend selecting a
better-validated judge before reporting a human-calibrated Elo scale.

\subsection{Conformal intervals on each corpus}
\label{app:conformal-cross-corpus}

The MAE reductions reported in Table~\ref{tab:extra-datasets} translate into
narrower conformal intervals on both extra corpora. We re-run
the same conformal pipeline used in Section~\ref{sec:conformal} on each
replication corpus, with no change to $\alpha = 0.10$, the normalised
nonconformity score $S_i = |\varepsilon_i| / \widehat{SE}_i$, or the
calibration/test split protocol; the only inputs that change are the underlying
Hard- and Soft-Elo BT fits.

Table~\ref{tab:cross-corpus-conformal} gives the cross-judge aggregate. On
LMArena 140K, Soft-Elo narrows conformal intervals for all reported
judge--corpus combinations, with reductions of $30$--$73\%$ (mean $56\%$). On
ComparIA, Soft-Elo again narrows intervals for all reported combinations, with
reductions of $19$--$78\%$ (mean $56\%$). The smallest ComparIA reductions are
on the Qwen judges, matching the low-$\bstar$ warning in
Appendix~\ref{app:comparia}. The bracketed ranges in the table show variation
across five random calibration/test splits.

\begin{table}[t]
\centering
\small
\caption{\textbf{Cross-corpus conformal payoff across random model-level splits.}
Entries report the mean median conformal interval width across five
calibration/test splits; brackets show the range across those splits. Soft-Elo
narrows intervals on every reported judge--corpus combination without changing
the conformal procedure.}
\label{tab:cross-corpus-conformal}
\setlength{\tabcolsep}{3pt}
\begin{tabular}{lll lr}
\toprule
Corpus & Judge & Hard width & Soft width & Reduction \\
\midrule
LMArena 140K & Gemma4-E4B     & 405 [394--417] & 110 [81--134]  & 73\% \\
LMArena 140K & Gemma4-26B-A4B & 405 [297--451] & 114 [93--122]  & 72\% \\
LMArena 140K & Qwen3.5-27B    & 361 [334--395] & 112 [89--132]  & 69\% \\
LMArena 140K & GPT-OSS-20B    & 268 [215--301] & 126 [98--182]  & 53\% \\
LMArena 140K & GPT-OSS-120B   & 230 [190--254] & 140 [122--177] & 39\% \\
LMArena 140K & Llama-3.3-70B  & 224 [172--281] & 158 [103--197] & 30\% \\
\midrule
ComparIA & Gemma4-26B-A4B & 622 [610--635] & 136 [115--154] & 78\% \\
ComparIA & Gemma4-E4B     & 545 [477--595] & 138 [120--153] & 75\% \\
ComparIA & GPT-OSS-20B    & 449 [406--471] & 130 [126--133] & 71\% \\
ComparIA & GPT-OSS-120B   & 399 [386--415] & 143 [138--152] & 64\% \\
ComparIA & Llama-3.3-70B  & 278 [263--314] & 147 [133--165] & 47\% \\
ComparIA & Qwen3.5-27B    & 339 [322--357] & 217 [196--249] & 36\% \\
ComparIA & Qwen3-32B      & 281 [235--308] & 229 [214--252] & 19\% \\
\bottomrule
\end{tabular}
\end{table}

\begin{figure}[t]
\centering
\includegraphics[width=\linewidth]{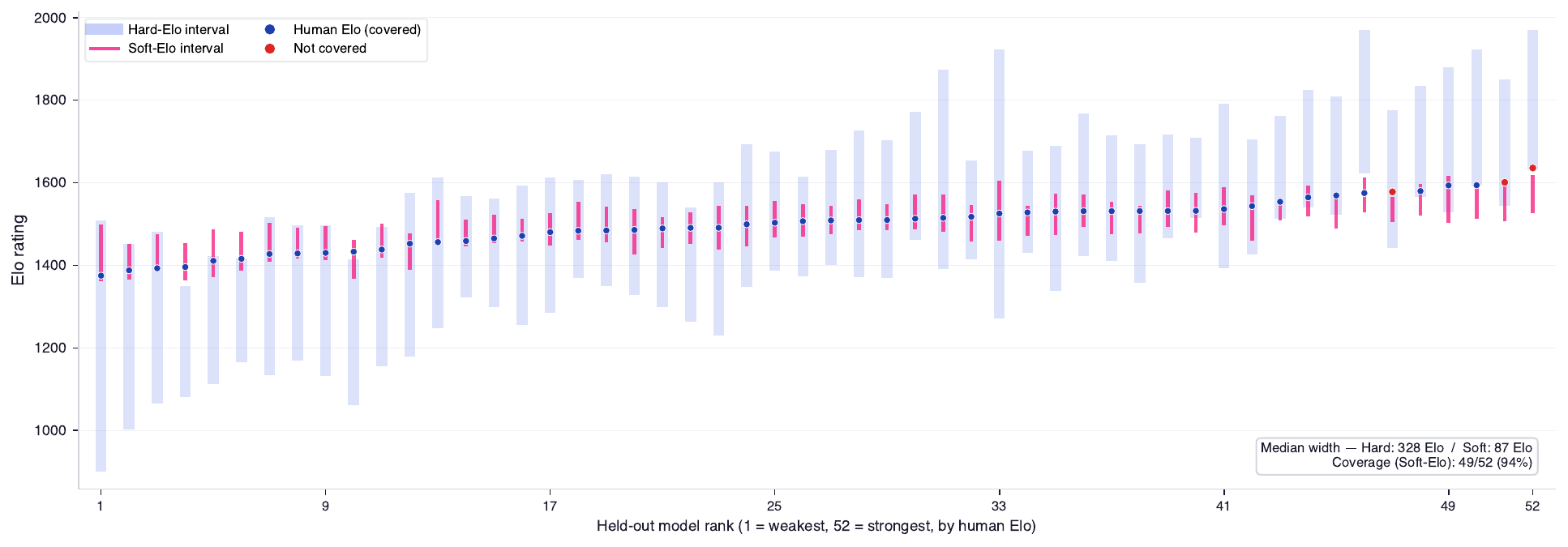}
\caption{\textbf{LMArena 140K: Soft-Elo contracts conformal intervals
by ${\sim}3.8\times$.} Per-model $90\%$ split-conformal intervals on the
$52$ held-out models, sorted by human Elo. Hard-Elo intervals (translucent
blue halos) reach above $500$~Elo at the strength tails; Soft-Elo
intervals (pink spines) shrink uniformly at similar empirical coverage.
Median width: Hard $328$~Elo, Soft $87$~Elo; coverage $49 / 52 = 94\%$.
Judge: Gemma4-26B-A4B.}
\label{fig:conformal-lmsys140k}
\end{figure}

\begin{figure}[t]
\centering
\includegraphics[width=\linewidth]{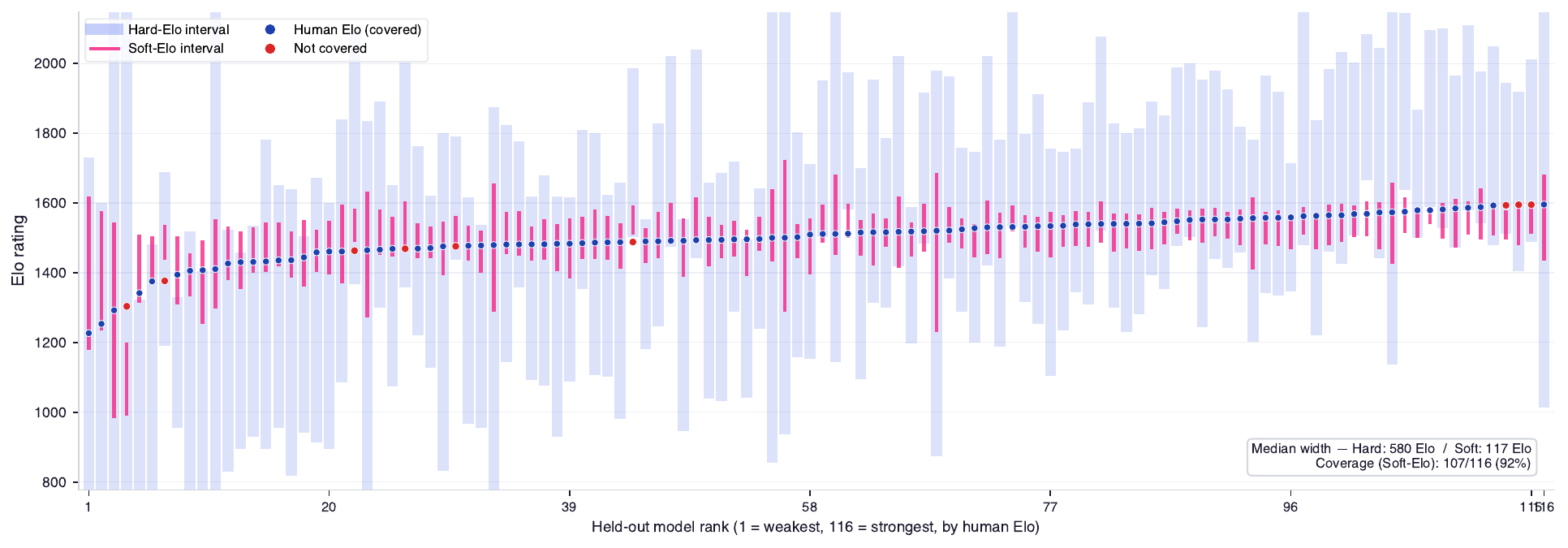}
\caption{\textbf{ComparIA: Soft-Elo narrows conformal intervals
by ${\sim}5.0\times$.} Same plot type as
Figure~\ref{fig:conformal-lmsys140k} on the $116$-model ComparIA pool with
Gemma4-26B-A4B as the judge, where the score difference retains its agreement
signal under the language shift. Median width: Hard $580$~Elo, Soft
$117$~Elo (${\sim}5.0\times$ contraction); coverage $107 / 116 = 92\%$.}
\label{fig:conformal-comparia}
\end{figure}

Figures~\ref{fig:conformal-lmsys140k} and~\ref{fig:conformal-comparia}
show per-model intervals for Gemma4-26B-A4B. At comparable empirical coverage,
the median interval contracts by $3.8\times$ on LMArena 140K and $5.0\times$
on ComparIA.

\subsection{Transferring calibration across corpora}
\label{app:deployment-shift}
Across five non-Qwen judges, transferring $\beta$ from LMArena 100K changes
mean Soft-Elo MAE from $23.0$ to $23.9$ on LMArena 140K and from $32.0$ to
$32.3$ on ComparIA, relative to fitting on the target corpus. These tests
exclude the Qwen failure cases above.

Freezing the source conformal quantile is more consequential: empirical
coverage falls to $71.2\%$ on LMArena 140K and $82.4\%$ on ComparIA.
Refreshing $\hat q$ from target-domain residuals raises coverage to
$94.0\%$ and $94.6\%$, respectively, with $\beta$ held fixed
(Table~\ref{tab:coverage_shift}). This requires human-reference information
for the target calibration models. A reliable transferred score map does
not by itself guarantee valid transferred intervals.
\begin{table}[htbp]
\centering
\small
\setlength{\tabcolsep}{4pt}
\caption{Coverage (\%) for nominal 90\% intervals transferred from LMArena 100K, with source $\beta$ fixed. Refreshing $\hat q$ uses target-domain calibration residuals. Brackets are 95\% judge-bootstrap intervals across five non-Qwen judges.}
\label{tab:coverage_shift}
\begin{tabular}{lrr}
\toprule
Target corpus & Source $\hat q$ & Refreshed $\hat q$ \\
\midrule
LMArena 140K & 71.2 [61.9, 81.2] & 94.0 [92.8, 95.2] \\
ComparIA & 82.4 [76.7, 88.1] & 94.6 [93.9, 95.4] \\
\bottomrule
\end{tabular}
\end{table}

\subsection{Models arriving later}
\label{app:temporal-arrival}
Using first-appearance timestamps in LMArena 140K, we evaluate three cutoffs
(May 1, 8, and 22, 2025). At each cutoff, temperature, conformal calibration,
and anchor ratings use only earlier models and battles. Each later model is
placed using its battles against the earlier models. The cutoffs provide
23--30 calibration models and 20--27 later models. Across five judges,
Soft-Elo attains mean coverage of $93.0\%$, compared with $77.8\%$ for
Hard-Elo, while reducing width and Elo MAE
(Table~\ref{tab:temporal_arrival}). These are empirical results for the tested
arrival sequence. Temporal or population shifts can still violate the
exchangeability assumption underlying the marginal coverage guarantee.
\begin{table}[htbp]
\centering
\small
\setlength{\tabcolsep}{4pt}
\caption{Chronological arrival evaluation: means across five judges and three cutoffs (May 1, 8, and 22, 2025). Coverage brackets show the minimum and maximum cutoff means. The coverage target is 90\%.}
\label{tab:temporal_arrival}
\begin{tabular}{lrrr}
\toprule
Method & Coverage (\%) & Median width (Elo) & Elo MAE \\
\midrule
Hard-Elo & 77.8 [66.0, 88.8] & 207.7 & 67.2 \\
Soft-Elo & 93.0 [89.6, 98.4] & 129.2 & 25.1 \\
\bottomrule
\end{tabular}
\end{table}

\subsection{Language subsets of LMArena}
\label{app:cross-lingual}

The main experiments use Chatbot Arena battles drawn from a multilingual prompt pool,
though the distribution is English-dominated ($13{,}423$ of $\approx 23{,}000$ shared
prompts, or $\approx 58\%$). We repeat the analysis across the 13 languages for
which all eight judges share at least 50 common prompts
($N_{\mathrm{prompts}} = 158$--$13{,}423$; Table~\ref{tab:lang-metrics}).

For each language subset we fit Hard-Elo and Soft-Elo BT independently using the
pooled $\bstar$ from the default calibration, then compare both to the global
human Elo reference.

\begin{figure}[h]
\centering
\includegraphics[width=\linewidth]{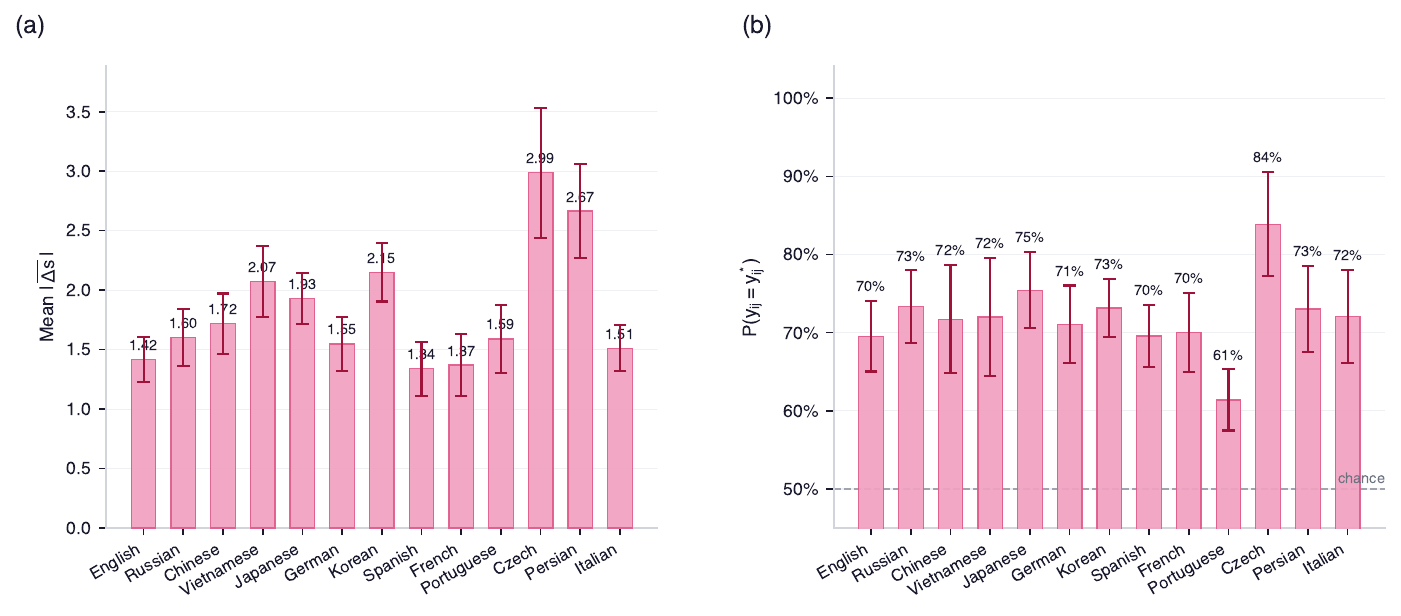}
\caption{\textbf{Cross-lingual signal: score difference and judge--human agreement.}
Languages sorted left to right by total decisive battle count.
(a)~Mean $|s|$ per language, averaged across the eight judges
(error bars: $\pm 1\,\mathrm{SD}$ across judges).
(b)~Cross-judge agreement rate $P(y_{ij} = y^*_{ij})$ on decisive battles, with the
chance level marked.}
\label{fig:cross-lingual}
\end{figure}

\paragraph{Score differences and judge--human agreement are stable across languages.}
Figure~\ref{fig:cross-lingual} shows two views over the 13 qualifying languages,
sorted left to right by total decisive battle count. Panel~(a): mean absolute score
difference $|s|$, averaged across the eight judges. Korean and Vietnamese produce larger mean score differences than English.
Panel~(b): cross-judge agreement rate $P(y_{ij} = y^*_{ij})$ on decisive battles.
This sits between $61\%$ and $84\%$ across languages, well above the chance level, with
the lowest-resource languages showing the noisiest estimates.

\paragraph{$\bstar$ is stable across high-resource languages.}
Fitting $\bstar$ separately on each language's decisive battles (using the same
$\beta$ MLE procedure as the default calibration; Section~\ref{sec:soft-method}) yields
per-judge values that stay within the global range $\bstar \in [0.36, 0.60]$ for the
seven high-resource languages (English through Korean, $N \geq 2{,}266$): the
cross-judge mean per-language $\bstar$ remains inside this band on every one of these
seven. For the lower-resource languages (Portuguese, Czech, Persian, Italian) the
cross-judge scatter grows, consistent with higher MLE variance at small battle
counts, and individual judges occasionally fall outside the band.

\paragraph{Rating accuracy by language.}
Table~\ref{tab:lang-metrics} summarises the hard vs.\ Soft-Elo comparison per language.
MAE falls by $74$--$85\%$ for every language. Spearman $\rho$ is stable:
the only negative change is Japanese ($\Delta\rho=-0.005$), while Korean
is essentially unchanged ($\Delta\rho=+0.005$); both are within estimation
noise at those battle counts. Hard-Elo MAE is
largest for low-resource languages (Persian $355$ Elo, Italian $334$ Elo) because sparse
battles allow extreme BT estimates; Soft-Elo corrects this most dramatically in absolute terms.

\begin{table}[t]
\centering\small\setlength{\tabcolsep}{5pt}
\caption{Per-language Hard-Elo vs.\ Soft-Elo performance, averaged across the eight judges. Languages are sorted by number of shared prompts $N_{\mathrm{prompts}}$ (unique instruction IDs present in all eight judge evaluation sets). $\Delta$MAE\% = relative MAE reduction; $\Delta\rho$ = Spearman $\rho$ gain. Soft-Elo reduces MAE for every language while leaving $\rho$ stable.}
\label{tab:lang-metrics}
\begin{tabular}{lrrrrrrrr}
\toprule
& & \multicolumn{3}{c}{Elo MAE $\downarrow$} & \multicolumn{3}{c}{Spearman $\rho$ $\uparrow$} \\
\cmidrule(lr){3-5}\cmidrule(lr){6-8}
Language & $N_{\mathrm{prompts}}$ & Hard-Elo & Soft-Elo & $\Delta$MAE\% & Hard-Elo & Soft-Elo & $\Delta\rho$ \\
\midrule
English & 13,423 & 86.6 & 20.6 & $\mathbf{76\%}$ & 0.920 & 0.923 & $+0.004$ \\
Russian & 2,737 & 129.7 & 27.9 & $\mathbf{79\%}$ & 0.878 & 0.885 & $+0.006$ \\
Chinese & 2,308 & 145.6 & 28.1 & $\mathbf{81\%}$ & 0.854 & 0.888 & $+0.034$ \\
Japanese & 903 & 170.2 & 37.3 & $\mathbf{78\%}$ & 0.807 & 0.801 & $-0.005$ \\
Vietnamese & 793 & 273.9 & 60.1 & $\mathbf{78\%}$ & 0.774 & 0.785 & $+0.011$ \\
German & 773 & 144.5 & 25.2 & $\mathbf{83\%}$ & 0.836 & 0.877 & $+0.042$ \\
Korean & 586 & 229.3 & 52.9 & $\mathbf{77\%}$ & 0.796 & 0.801 & $+0.005$ \\
Spanish & 418 & 197.2 & 35.6 & $\mathbf{82\%}$ & 0.766 & 0.786 & $+0.020$ \\
French & 331 & 231.6 & 35.7 & $\mathbf{85\%}$ & 0.666 & 0.736 & $+0.070$ \\
Portuguese & 285 & 269.2 & 41.3 & $\mathbf{85\%}$ & 0.645 & 0.745 & $+0.100$ \\
Czech & 217 & 308.3 & 80.8 & $\mathbf{74\%}$ & 0.661 & 0.691 & $+0.029$ \\
Italian & 175 & 333.6 & 58.1 & $\mathbf{83\%}$ & 0.507 & 0.596 & $+0.089$ \\
Persian & 158 & 354.8 & 90.6 & $\mathbf{74\%}$ & 0.680 & 0.778 & $+0.098$ \\
\bottomrule
\end{tabular}
\end{table}

\paragraph{Pooled $\bstar$ is sufficient.}
All results above use the pooled $\bstar$ from the default calibration.
We also fit a separate $\bstar$ on each language's decisive human-labeled
battles using the same MLE procedure (Section~\ref{sec:soft-method}) and
re-evaluate MAE.

For 10 of 13 languages the difference in MAE between pooled and per-language $\bstar$ is
within $\pm 5$~Elo, consistent with estimation noise rather than a systematic gap.
Among the sparse language subsets,
Czech ($N_{\mathrm{prompts}} = 217$) incurs $+21.8$~Elo \emph{higher} MAE under per-language
calibration (102.7 vs 80.8~Elo pooled), and Italian ($N = 175$) similarly worsens by
$3.6$~Elo. The per-language MLE overfits on sparse data, pushing $\bstar$ toward extreme
values that the pooled estimate avoids.
Only Portuguese ($N = 285$) shows a non-trivial improvement from per-language calibration
($+9.3$~Elo), within sampling variance at that budget.

We therefore recommend the pooled $\bstar$ as the default: it requires no
language-specific annotation, is more robust at low budgets, and performs
comparably to per-language calibration for all languages considered here.

\section{Annotation Budget and Sample Efficiency}
\label{app:annotation-budget}
\subsection{Temperature stability and sample efficiency}
\label{app:beta-stability}

\begin{figure}[h]
\centering
\includegraphics[width=\linewidth]{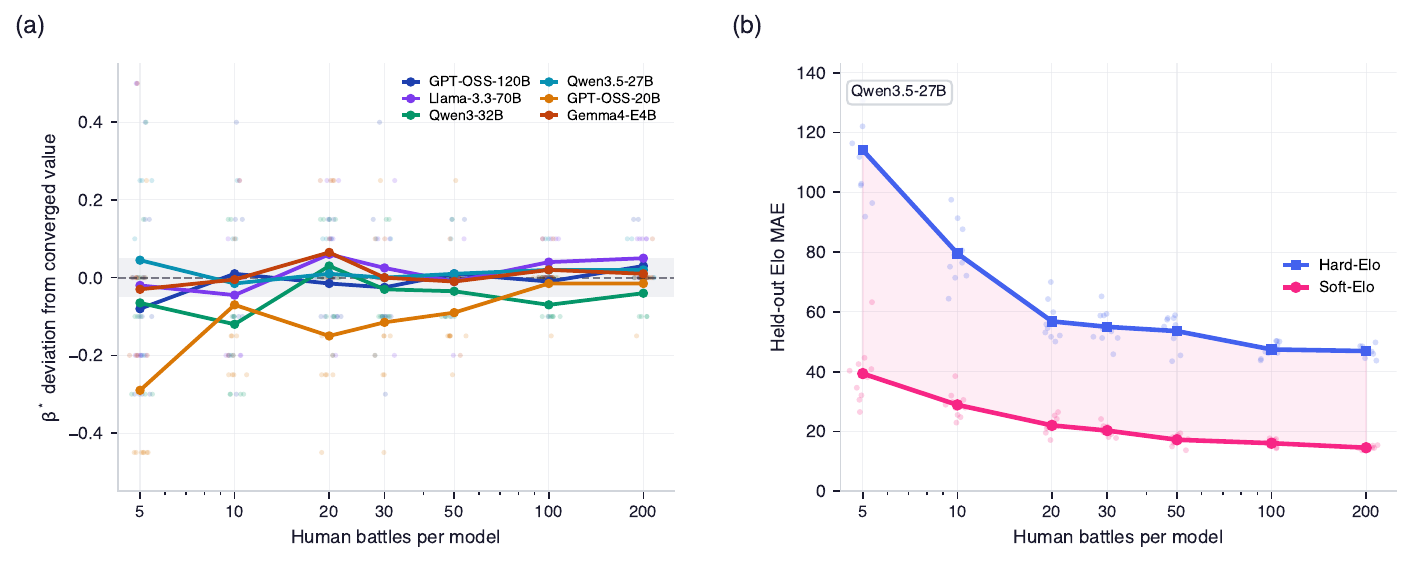}
\caption{\textbf{$\bstar$ stabilises rapidly and Soft-Elo beats Hard-Elo at every
annotation budget.}
(a)~Per-budget deviation of the selected $\bstar$ from each judge's converged
value (mode at $b = 200$). Across judges, $\bstar \in [0.36, 0.60]$, and the
per-judge mean lies within $\pm 0.05$ of its converged value by
$b \approx 50$. Coloured dots: random subsets per budget; coloured lines:
per-judge mean. Shaded band: $\pm 0.05$.
(b)~Sample efficiency on Qwen3.5-27B: Soft-Elo's held-out MAE drops from
$\approx 81$ to $\approx 17$~Elo as the annotation budget grows from
$b = 5$ to $b = 200$, while Hard-Elo only falls from $\approx 125$ to
$\approx 48$~Elo; the shaded gap marks the Soft-Elo advantage. The
cross-judge median pattern is similar: Soft-Elo MAE $86 \to 19$~Elo against
Hard-Elo $123 \to 47$~Elo over the same budget range.}
\label{fig:beta-stability}
\end{figure}

We vary two calibration budgets: battles per model $b$
(Appendix~\ref{app:judge-protocol}) and models used to fit $\bstar$.
Figure~\ref{fig:beta-stability} reports the effect of the battle budget and
compares temperatures fitted with and without each target model.
Table~\ref{tab:calibration_models} varies the number of calibration models
while holding the anchor and conformal procedures fixed.

Across all eight judges, $\bstar$ converges to $[0.36,0.60]$ and lies
within $\pm0.05$ of its converged value by $b\approx50$ battles per model
(Figure~\ref{fig:beta-stability}). Excluding each target model from calibration
changes the fitted temperature by $0.0026$ on average and at most $0.016$
relative to the pooled fit. The corresponding Soft-Elo MAE changes by
$0.09$~Elo on average and at most $0.18$~Elo across judges.
Soft-Elo has lower MAE than Hard-Elo at every tested budget, including $b=5$.

\subsection{Number of calibration models}
We hold out 11 models and vary the temperature-fitting pool from
5 to 44 models over 12 draws for three judges. The anchor and conformal
procedures remain fixed (Table~\ref{tab:calibration_models}). Five calibration
models retain most of the MAE improvement in this experiment; increasing the
pool from 5 to 44 models changes MAE from $19.2$ to $17.8$ Elo. These counts
concern temperature fitting only, not the complete human-label budget needed
for the anchor leaderboard and conformal calibration. The battle-budget
experiment in Section~\ref{app:beta-stability} varies a separate axis;
together these experiments do not establish a joint minimum budget.
\begin{table}[htbp]
\centering
\small
\setlength{\tabcolsep}{4pt}
\caption{Varying only the number of models used to fit $\beta$. Entries average three judges and 12 draws; the anchor and conformal procedures are held fixed. Hard-Elo gives MAE 50.1, $\rho=0.951$, mean width 211.4 Elo, and coverage 92.0\%.}
\label{tab:calibration_models}
\begin{tabular}{rrrrr}
\toprule
Models for $\beta$ & Elo MAE & $\rho$ & Mean width (Elo) & Coverage (\%) \\
\midrule
5 & 19.2 & 0.941 & 88.6 & 92.3 \\
10 & 18.3 & 0.950 & 83.5 & 92.2 \\
20 & 18.0 & 0.952 & 82.8 & 92.3 \\
30 & 17.9 & 0.953 & 81.8 & 92.2 \\
44 & 17.8 & 0.953 & 82.2 & 92.3 \\
\bottomrule
\end{tabular}
\end{table}

\section{Calibration and Reference Uncertainty}
\label{app:calibration-sensitivity}

We check whether fitted probabilities match observed human agreement and
how changes to the scoring prompt affect Elo estimates. We then test whether
Soft-Elo's advantage persists when the human reference leaderboard is
re-estimated from resampled battles.

\subsection{Probability calibration}
\label{app:calibration}

Section~\ref{sec:soft-method} treats $\sigma(\bstar s(x))$ as a calibrated
probability that $A$ beats $B$. To evaluate calibration as judge confidence,
we map both presentation orders to the same model identities and average their
score gaps. On battles where both human and judge select a winner, we orient
the probability toward the judge's selected side:
\[
  p_{\mathrm{corr}}(x)=\sigma(\bstar |s(x)|),
\]
the predicted probability that the judge's chosen side matches the human
choice. We bin $p_{\mathrm{corr}}(x)$ into ten equal-mass deciles and compare
each bin's mean prediction to the empirical agreement
$\mathbf{1}[y_{ij}(x)=y^*_{ij}(x)]$, reporting the Expected Calibration Error
(ECE). Writing $p_i=p_{\mathrm{corr}}(x_i)$ for battle $i$, $n$ for
the number of battles, and $n_b$ for the size of bin $b$,
\[
 \mathrm{ECE}=\sum_b\frac{n_b}{n}
 \left|\frac{1}{n_b}\sum_{i\in b}p_i
 -\frac{1}{n_b}\sum_{i\in b}\mathbf{1}[y_i=y_i^*]\right|.
\]

\begin{figure}[h]
\centering
\includegraphics[width=\linewidth]{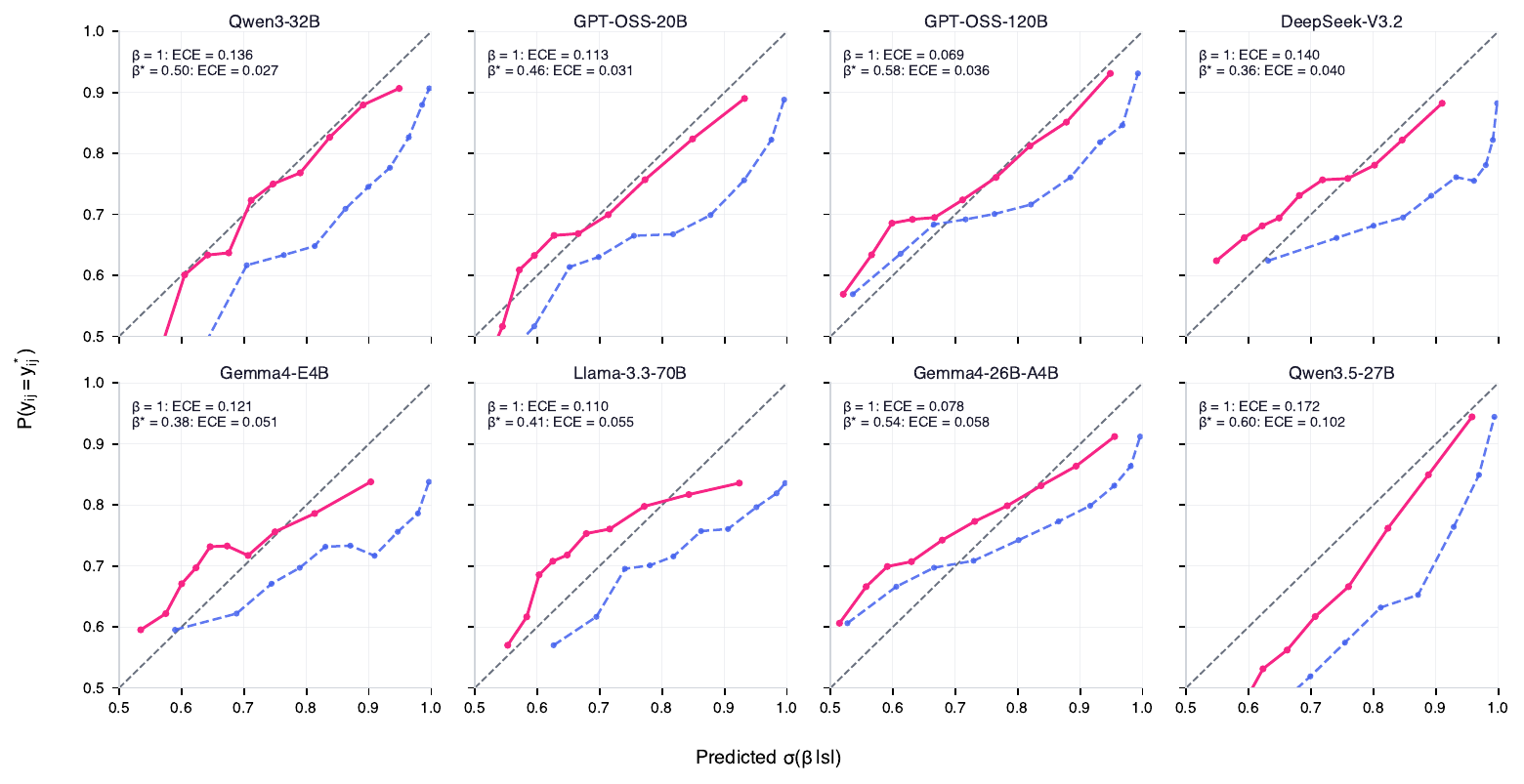}
\caption{\textbf{Per-judge calibration of $\sigma(\bstar |s|)$ as
the probability that the judge's chosen side matches the human choice.}
Each panel shows empirical
agreement vs.\ predicted probability across ten equal-mass deciles;
the diagonal marks perfect calibration. Seven of eight judges are
well-calibrated ($\text{ECE} \le 0.06$); Qwen3.5-27B is systematically
over-confident ($\text{ECE} = 0.10$). Panels sorted by ECE.}
\label{fig:margin-calibration}
\end{figure}

Figure~\ref{fig:margin-calibration} shows the per-judge reliability
diagrams. For seven of the eight judges, ECE falls in
$[0.027, 0.058]$ and the curves track the diagonal closely across
predicted probabilities in $[0.5, 0.95]$. The mean ECE across all
judges is $0.050$ and the median is $0.045$. Qwen3.5-27B is
the worst-calibrated case ($\text{ECE} = 0.10$): its empirical
accuracy sits roughly $10$ percentage points below the predicted
probability uniformly across bins, indicating systematic
over-confidence. This is also the judge with the highest $\bstar$
in our set ($0.60$) and the lowest empirical agreement ($60.8\%$):
its score differences are wide in absolute units but less reliable than
the magnitude suggests.

\paragraph{Why calibrate $\bstar$ on human non-ties?}
The temperature MLE in Eq.~\ref{eq:temp-mle} estimates a binary preference map:
given a signed score difference $s(x)$, it asks how likely humans are to prefer
$A$ over $B$. Human ties do not give a clean orientation for that binary map.
As a sensitivity check, we also fit $\bstar$ on all human-labeled battles by
treating human ties as target $0.5$. This all-battle fit is deliberately more
conservative: it roughly halves $\bstar$ relative to the non-tie fit
($\beta_{\mathrm{all}}/\beta_{\mathrm{non\text{-}tie}}\approx 0.56$ in our
sensitivity run), leaves rank correlation essentially unchanged
($\Delta\rho\approx -0.002$), and increases held-out Elo MAE by about
$7$~Elo on average. We therefore use human non-tie comparisons to calibrate
the binary preference map, then apply the resulting soft target to every
judge-scored battle, including near-ties.

\begin{figure}[h]
\centering
\includegraphics[width=0.55\linewidth]{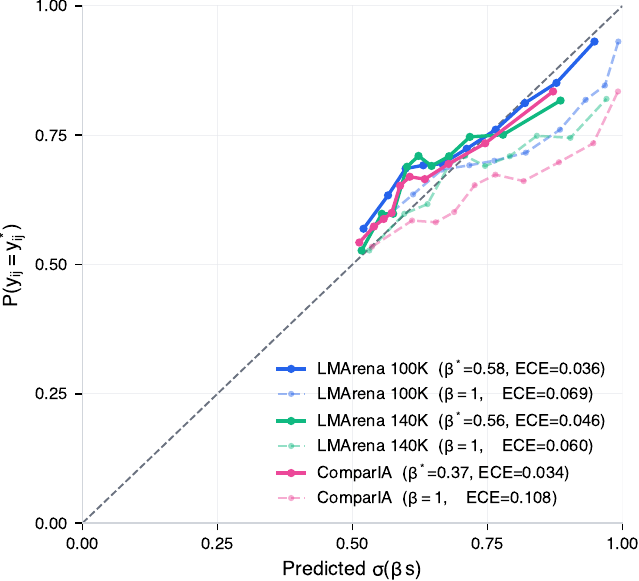}
\caption{\textbf{Calibration of $\sigma(\bstar\,|s|)$ for GPT-OSS-120B
with a separate temperature fit on each corpus; ECE remains $\le 0.07$.}
Reliability diagram for GPT-OSS-120B on LMArena 100K, LMArena 140K, and
ComparIA. Per-corpus $\bstar$ and ECE in the legend; dashed line is
perfect calibration.}
\label{fig:calibration-corpora}
\end{figure}

\paragraph{Cross-corpus calibration.}
We repeat the reliability analysis on LMArena 140K and ComparIA, refitting
$\bstar$ for each judge--corpus pair (Table~\ref{tab:calibration-corpora}).
Figure~\ref{fig:calibration-corpora} compares the reliability curves for
GPT-OSS-120B across corpora.
Calibration remains close on LMArena 140K and on 5 of 7 ComparIA judges. Both Qwen judges on ComparIA fall outside
the well-calibrated regime ($\text{ECE} = 0.080$ and $0.101$,
$\bstar = 0.14$ and $0.13$): the same bifurcation flagged in
Appendix~\ref{app:comparia}.

\begin{table}[h]
\centering
\small
\setlength{\tabcolsep}{4pt}
\caption{Per-judge calibration of $\sigma(\bstar\,s)$ as a probability of judge correctness, across the three corpora. Each cell shows ECE / $\bstar$. Empty cells: judge not available in that corpus. \textbf{Bold}: ECE worse than $0.07$ (calibration breakdown).}
\label{tab:calibration-corpora}
\begin{tabular}{lcccccc}
\toprule
& \multicolumn{2}{c}{LMArena 100K} & \multicolumn{2}{c}{LMArena 140K} & \multicolumn{2}{c}{ComparIA} \\
\cmidrule(lr){2-3}\cmidrule(lr){4-5}\cmidrule(lr){6-7}
Judge & ECE & $\bstar$ & ECE & $\bstar$ & ECE & $\bstar$ \\
\midrule
Qwen3-32B      & 0.027 & 0.50 & --- & --- & \textbf{0.101} & 0.13 \\
Qwen3.5-27B    & \textbf{0.102} & 0.60 & 0.040 & 0.51 & \textbf{0.080} & 0.14 \\
GPT-OSS-20B    & 0.031 & 0.46 & 0.053 & 0.51 & 0.044 & 0.34 \\
GPT-OSS-120B   & 0.036 & 0.58 & 0.046 & 0.56 & 0.034 & 0.37 \\
DeepSeek-V3.2  & 0.040 & 0.36 & 0.051 & 0.40 & ---            & ---  \\
Gemma4-E4B     & 0.051 & 0.38 & 0.044 & 0.41 & 0.029 & 0.30 \\
Gemma4-26B-A4B & 0.058 & 0.54 & \textbf{0.091} & 0.60 & 0.057 & 0.34 \\
Llama-3.3-70B  & 0.055 & 0.41 & 0.057 & 0.47 & 0.063 & 0.32 \\
\midrule
Mean ECE       & \multicolumn{2}{c}{0.050} & \multicolumn{2}{c}{0.054} & \multicolumn{2}{c}{0.058} \\
\bottomrule
\end{tabular}
\end{table}

These reliability diagrams support approximate calibration for most tested
judge--corpus pairs. Residual miscalibration remains, especially for Qwen on
ComparIA (Appendix~\ref{app:comparia}).

We additionally observe that $11.6\%$ of decisive battles exhibit a
verdict-vs-score sign disagreement, where the judge labels the
side it scored lower. This is computed at the per-battle
level and is independent of the human-judge agreement question; it
reflects internal inconsistency in the judge's output that the soft
target absorbs through the signed-score-difference formulation.

\subsection{Score range and prompt sensitivity}
\label{app:prompt-budget}

For a positive affine rescaling $s'(x)=a s(x)$, setting
$\beta'=\beta/a$ leaves the target unchanged:
\[
 \sigma(\beta' s'(x))=\sigma(\beta s(x)).
\]
In particular, normalizing scores from $[m,M]$ to $[0,1]$ rescales gaps by
$1/(M-m)$ and can be absorbed by refitting $\beta'=(M-m)\beta$.
If observed minima and maxima are used, they must be estimated on the
calibration fold. Fixing $\beta=1$ after normalization instead imposes an
uncalibrated temperature.

Actual prompts can change how judges use a scale. We therefore re-judge a
5,000-battle sample spanning the evaluation model pool with Qwen3-32B,
Gemma4-26B-A4B, and
Llama-3.3-70B. We vary the six-criterion score range and also test one overall
score. After parsing and filtering, 4,493--5,000 battles remain per condition.
We fit one pooled $\beta$ per judge and format before held-out Elo placement
(Table~\ref{tab:prompt_scale}). Because the temperature fit includes each
target model, this diagnostic does not test calibration on unseen models.

\begin{table}[htbp]
\centering
\small
\setlength{\tabcolsep}{4pt}
\caption{Pairwise prompt sensitivity: Hard / Soft Elo MAE. The four numerical ranges use six criteria; Overall uses one score on 0--10. This diagnostic fits one pooled temperature per judge and format, then holds out models for Elo placement.}
\label{tab:prompt_scale}
\begin{tabular}{lrrrrr}
\toprule
Judge & 0--1 & 0--5 & 1--10 & 0--100 & Overall \\
\midrule
Qwen3-32B & 59.1 / 18.9 & 56.4 / 19.3 & 31.2 / 22.1 & 48.0 / 21.5 & 40.5 / 21.3 \\
Gemma4-26B-A4B & 70.0 / 19.1 & 69.8 / 18.8 & 55.6 / 19.3 & 61.3 / 20.2 & 61.2 / 19.4 \\
Llama-3.3-70B & 53.5 / 29.5 & 39.6 / 28.1 & 44.8 / 29.1 & 41.7 / 29.6 & 46.7 / 28.7 \\
$\beta$ range & 4.423--4.892 & 0.785--1.047 & 0.419--0.518 & 0.037--0.051 & 0.283--0.427 \\
\bottomrule
\end{tabular}
\end{table}

Within this diagnostic, Soft-Elo lowers MAE in all 15 combinations. Its
within-judge MAE range is $1.4$--$3.1$ Elo, compared with $13.9$--$27.9$ for
Hard-Elo. Refitting $\beta$ absorbs much of the score-range change, but the
invariance argument alone does not guarantee robustness to arbitrary prompts
or to judges whose score gaps are uninformative.

\subsection{Uncertainty in the human reference}
\label{app:human-reference}
We hold judge-derived estimates fixed and evaluate them against 200 bootstrap
refits of the human leaderboard on the shared battles
(Appendix~\ref{app:shared-battles})
(Table~\ref{tab:human_reference_robustness}). This measures finite-sample
variation in the reference; each source battle has one human preference,
so the analysis does not measure inter-annotator disagreement.
Soft-Elo has lower mean MAE in all 200 refits. The mean reduction is
$27.25$ Elo, with a central 95\% range of $[23.26,30.83]$.
Mean rank correlations are $0.920$ for Hard-Elo and $0.926$ for Soft-Elo;
we do not claim a significant ordering improvement.
\begin{table}[htbp]
\centering
\small
\setlength{\tabcolsep}{4pt}
\caption{Fixed judge-based estimates evaluated against the full human reference and 200 bootstrap refits of that reference. Brackets are central 95\% ranges; these quantify finite-sample uncertainty, not inter-annotator disagreement.}
\label{tab:human_reference_robustness}
\begin{tabular}{lrr}
\toprule
Human reference & Hard-Elo MAE & Soft-Elo MAE \\
\midrule
Full leaderboard & 47.74 & 18.70 \\
Bootstrap refits & 48.14 [45.43, 51.11] & 20.90 [18.70, 22.89] \\
\bottomrule
\end{tabular}
\end{table}

\section{Label Smoothing and Judge Committees}
\label{app:baselines}
\subsection{Label smoothing}
\label{app:label-smoothing}
\newcommand{\LSBestC}{0.75}
\newcommand{\LSBestMAE}{17.5}
\newcommand{\LSSoftMAE}{17.8}
\newcommand{\LSHighCMAE}{35.4}

We compare score-dependent soft targets with constant label smoothing to
separate the use of score magnitudes from the effect of moving targets away
from $\{0,1\}$. This retrospective ablation holds out each model for Elo
placement but uses one pooled temperature per judge for Soft-Elo. It does
not test temperature calibration on unseen models.

\paragraph{Constant soft targets.}

Label smoothing replaces each decisive battle's hard label $y \in \{0, 1\}$
with a constant target $\tilde y_c \in \{c, 1{-}c\}$ for $c \in (0.5, 1)$.
Tie battles are handled by the same swap-consistency convention used
elsewhere in the paper. This is a special case of Soft-Elo: setting
$\beta = \mathrm{logit}(c)$ and replacing $s$ by its
sign $\mathrm{sgn}(s)$ in Eq.~\ref{eq:soft-target}
recovers the constant-$c$ target exactly. Soft-Elo strictly contains
label smoothing; the empirical question is how much the score-difference
\emph{magnitude} contributes beyond the sign.

\paragraph{Rating accuracy.}

We sweep $c \in \{0.55, 0.65, 0.75, 0.85, 0.95\}$, refitting the anchor
leaderboard and placing each held-out model for every value.
Table~\ref{tab:label-smoothing} reports the resulting MAEs alongside the
Hard-Elo and pooled-temperature Soft-Elo baselines from the same ablation.
The cross-judge mean is U-shaped in
$c$: $c = 0.55$ produces near-degenerate targets and high MAE,
$c = 0.95$ rises back toward Hard-Elo, and the minimum sits at
$c^* = \LSBestC{}$ with mean MAE $\LSBestMAE{}$~Elo, compared with
$\LSSoftMAE{}$~Elo for Soft-Elo in this ablation.
This comparison also reflects different fitting objectives: $\beta$ is
estimated by likelihood on human preferences, whereas the best smoothing
constant is selected retrospectively to minimize held-out Elo MAE.
The per-judge optimum sits at $0.75$ or $0.85$; no single
constant is optimal across the eight judges. At $c = 0.95$, mean MAE is
$\LSHighCMAE{}$~Elo.

Unlike constant smoothing, Soft-Elo varies its target continuously with the
score difference, distinguishing narrow from decisive preferences.

\begin{table}[h]
\centering
\small
\caption{Per-judge held-out Elo MAE in the constant-smoothing ablation.
Soft-Elo uses one pooled temperature per judge across all held-out models,
rather than the fold-specific calibration described in
Section~\ref{sec:soft-method}. All entries come from this ablation;
means are computed from unrounded per-judge MAEs.
The bold cell in each row marks the best constant in the retrospective sweep.}
\label{tab:label-smoothing}
\setlength{\tabcolsep}{4pt}
\begin{tabular}{lrrrrrrr}
\toprule
 & & \multicolumn{5}{c}{Label smoothing $c$} & \\
\cmidrule(lr){3-7}
Judge & Hard-Elo & 0.55 & 0.65 & 0.75 & 0.85 & 0.95 & Soft-Elo \\
\midrule
DeepSeek-V3.2 & 63.4 & 39.5 & 24.8 & \textbf{16.1} & 26.1 & 49.5 & 17.1 \\
GPT-OSS-120B & 47.4 & 40.4 & 25.5 & \textbf{13.7} & 16.6 & 36.2 & 14.4 \\
GPT-OSS-20B & 34.5 & 41.3 & 27.9 & 18.2 & \textbf{15.7} & 25.7 & 19.8 \\
Gemma4-26B-A4B & 55.9 & 39.9 & 23.6 & \textbf{12.2} & 20.5 & 42.9 & 15.6 \\
Gemma4-E4B & 48.2 & 40.4 & 27.6 & \textbf{20.6} & 22.1 & 37.5 & 20.8 \\
Llama-3.3-70B & 43.9 & 41.4 & 30.1 & \textbf{24.7} & 24.8 & 36.3 & 24.4 \\
Qwen3-32B & 27.5 & 42.0 & 29.1 & 18.7 & \textbf{13.6} & 20.4 & 16.7 \\
Qwen3.5-27B & 46.0 & 42.0 & 27.7 & \textbf{15.9} & 16.6 & 34.8 & 13.6 \\
\midrule
Mean & 45.8 & 40.9 & 27.0 & 17.5 & 19.5 & 35.4 & 17.8 \\
\bottomrule
\end{tabular}
\end{table}

\paragraph{Parameter fitting.}

The likelihood fit uses the parametric model
$P[B \prec A \mid x] = \sigma(\beta\,s)$.
The main procedure excludes the target model when fitting this map
(Section~\ref{sec:soft-method}); the present ablation reuses the pooled fit.
The retrospectively selected $c$ uses information unavailable when evaluating a new model
without human labels.

Score-gap diagnostics can also be checked on human-labeled calibration
battles before deployment (Appendix~\ref{app:comparia}). The retrospectively
selected smoothing constant does not provide that check.

\subsection{Judge committees}
\label{app:committees}

We compare single judges with committees on the shared battles.
The single-judge conditions average metrics from eight
separate leaderboards. The committee conditions fit one leaderboard from
all eight judgments per battle:
\begin{itemize}[nosep]
\item \textbf{Categorical mode:} use the most frequent A-win, tie, or B-win
label; a tie in category counts produces a tie target. This retains the tie
category, unlike taking a majority only among decisive A/B votes.
\item \textbf{Vote share:} average categorical targets encoded as
$1$, $\tfrac12$, and $0$.
\item \textbf{Calibrated scores:} average the eight judge-specific
probabilities, $\bar p(x)=\tfrac18\sum_{r=1}^{8}\sigma(\beta_r s_r(x))$,
with judged ties assigned $\tfrac12$ under this ablation's tie convention.
\end{itemize}
We use five repeats of five strength-balanced model folds. Each target is
excluded from both temperature fitting and anchor fitting. The score-based
committee uses the fold-specific judge temperatures; it does not fit a new
committee temperature. For a common uncertainty comparison, all conditions
use unnormalized absolute-residual conformal scores, with 38 calibration
and 17 test models over 500 splits. The main experiment instead uses
normalized scores; its widths are not directly comparable to this ablation.

\begin{table}[htbp]
\centering
\small
\setlength{\tabcolsep}{4pt}
\caption{Committee ablation on 18,158 shared battles and 55 models. MAE and rank-correlation brackets are 95\% model-bootstrap intervals; width brackets are the central 95\% range over 500 splits. All conditions use absolute-residual conformal intervals with a 90\% target. Single-judge rows average eight estimators.}
\label{tab:committee_comparison}
\begin{tabular}{lrrr}
\toprule
Estimator & Elo MAE & $\rho$ & Mean width (Elo) \\
\midrule
Single Hard-Elo & 47.58 [40.70, 54.47] & 0.936 [0.886, 0.960] & 212.7 [187.7, 227.2] \\
Single Soft-Elo & 18.50 [15.68, 21.83] & 0.942 [0.897, 0.963] & 87.4 [71.8, 104.3] \\
Committee categorical mode & 55.95 [46.84, 65.55] & 0.958 [0.910, 0.979] & 236.6 [194.2, 252.1] \\
Committee vote-share & 43.72 [36.38, 51.12] & 0.958 [0.908, 0.979] & 188.1 [174.0, 194.6] \\
Committee calibrated scores & 15.93 [12.43, 19.58] & 0.963 [0.922, 0.981] & 79.1 [56.9, 98.5] \\
\bottomrule
\end{tabular}
\end{table}

The calibrated-score committee reduces MAE to $15.93$ Elo, compared with
$18.50$ for the average single Soft-Elo judge and $43.72$ for vote share.
The categorical-mode committee has high rank correlation but retains a
substantial scale error ($55.95$ Elo). This supports using calibrated scores
within a committee, at the cost of eight judge evaluations per battle.
It does not establish superiority over every individual judge or reproduce
the full Auto-Arena pipeline.

\section{Judge Biases and Residual Structure}
\label{app:residual-structure}

We examine position, verbosity, and same-family effects as possible sources of
the residual pattern in Section~\ref{sec:anatomy}, then compare residuals
across model-strength quartiles after applying Soft-Elo.

\subsection{Battle-level biases: position, verbosity, and self-preference}
\label{app:bias-details}

\begin{figure}[t]
\centering
\includegraphics[width=\linewidth]{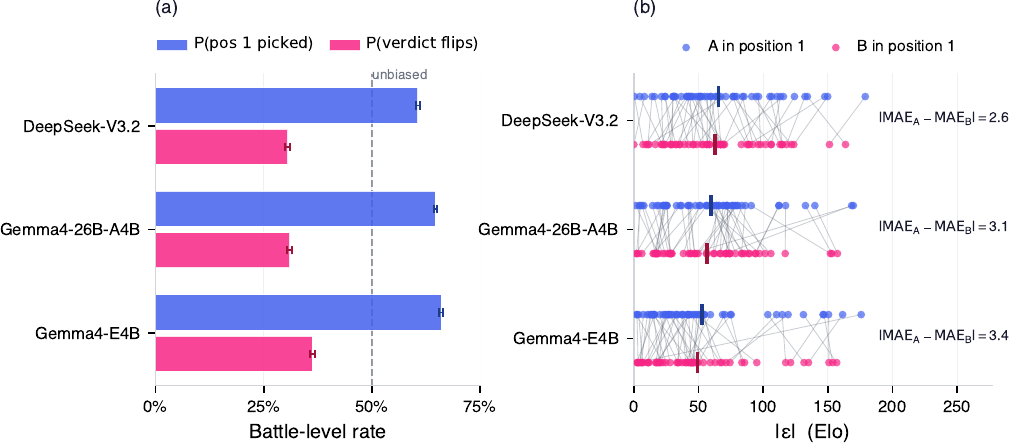}
\caption{\textbf{Position bias on the three judges with full swap evaluation.}
(a)~Battle-level rates: $P(\text{judge picks position 1})$ pooled
over both presentations (blue) and $P(\text{verdict flips when
positions swap})$ on decisive battles (pink); error bars are
$95\%$ confidence intervals.
(b)~Per-model $|\varepsilon|$ under Hard-Elo fitted on each
presentation independently: A in position 1 (blue, top lane) and
B in position 1 (pink, bottom lane). Each held-out model contributes
one dot per lane; grey connectors link a model's two values. Thick
coloured ticks mark the per-judge MAE; the right margin shows the
aggregate spread $|\mathrm{MAE}_A - \mathrm{MAE}_B|$.}
\label{fig:position-bias}
\end{figure}

\paragraph{Position bias.}
\label{app:position-bias}
At the battle level (Figure~\ref{fig:position-bias}a), every judge
with full per-presentation data prefers the position-1 response,
with $P(\text{pos\,1 picked}) = 61$--$66\%$ across the three
judges. On $30$--$36\%$ of decisive battles the verdict
\emph{flips} between the two presentations, so a near-third of the
binary outcome mass is order-dependent.

The same effect is present in the underlying per-completion scores:
averaged across both presentations, the position-1 response receives
a mean score advantage of around $+0.5$ points on the $1$--$10$
scale, with $69$--$78\%$ of battles directionally favouring position 1. The score difference $s$
therefore inherits the bias. All experiments fit both
Hard-Elo and Soft-Elo on swap-averaged battles rather than on either
presentation alone.

At the Elo level (Figure~\ref{fig:position-bias}b), per-judge
Hard-Elo MAE differs by $|\mathrm{MAE}_A - \mathrm{MAE}_B| \approx
3$~Elo between the two single-order fits, with per-model deflections
reaching $100+$~Elo on individual frontier models. Swap-averaged
fitting sits between the two extremes, reducing but not eliminating position
bias before residuals propagate downstream. The observed aggregate effect is small relative to total
Hard-Elo MAE (Table~\ref{tab:judges}), although individual frontier models
can still move substantially under a single presentation order.

\begin{figure}[t]
\centering
\includegraphics[width=0.65\linewidth]{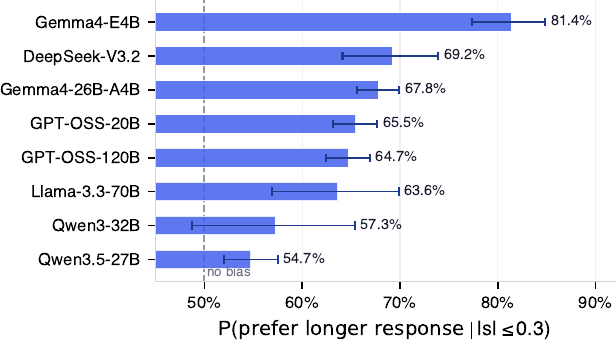}
\caption{\textbf{Verbosity bias.}
$P(\text{judge prefers longer response})$ on battles where the
judge's own score difference is small ($|s| \le 0.3$); error bars
are $95\%$ confidence intervals. Conditioning on near-tied judge scores reduces measured quality differences;
deviations from $50\%$ suggest a remaining length preference. The effect ranges
from $55\%$ (Qwen judges) to $81\%$ (Gemma4-E4B).}
\label{fig:verbosity-bias}
\end{figure}

\paragraph{Verbosity bias.}
\label{app:verbosity-bias}
At the battle level (Figure~\ref{fig:verbosity-bias}), every judge
prefers the longer response on battles its own scores call near-tied
($|s| \le 0.3$). Conditioning on near-tied judge scores reduces the role of
quality differences under the judge's own scoring rule, so deviations from
$50\%$ suggest a remaining length premium. The effect is universal but uneven,
ranging from $55\%$ (Qwen judges) to $81\%$ (Gemma4-E4B).

At the residual level, the per-(judge, model) Hard-Elo residual
correlates with mean response length even after partialling out
human Elo on three of the eight judges, with partial correlations
in the range $r \approx 0.3$--$0.6$. On the other five the partial
correlation is small or non-significant once strength is
controlled, consistent with length acting as a strength covariate
rather than an independent bias channel.

\begin{figure}[t]
\centering
\includegraphics[width=\linewidth]{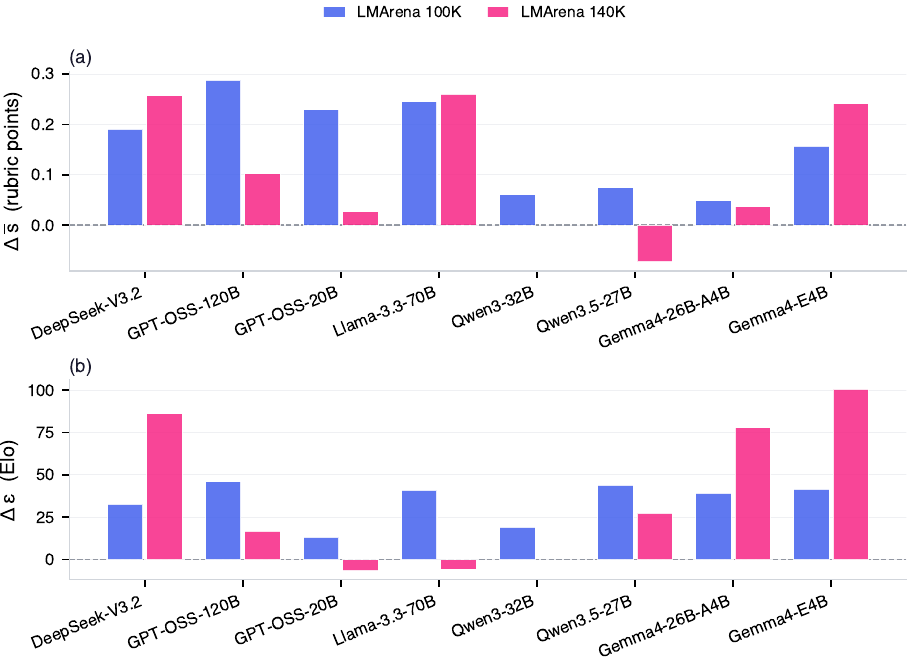}
\caption{\textbf{Same-family effects at the score and Elo-residual levels,
on LMArena 100K and 140K.}
Panel~(a) shows the same-family minus cross-family excess in raw judge scores
relative to the cross-judge median for each target model. Panel~(b) shows the
analogous excess in Elo residuals. Positive values indicate that a judge gives
more credit to models from its own family than to cross-family models,
relative to other judges.}
\label{fig:self-preference}
\end{figure}

\paragraph{Same-family effects.}
\label{app:self-preference}
We test whether each judge gives more credit to models from its own family
relative to how the other judges score the same models. For each judge $j$
and held-out target $m$, we measure $j$'s mean per-completion score (on the
$1$--$10$ scale) for $m$ and subtract the median across the other judges on
the same $m$; this gives a per-target deviation from cross-judge consensus.
We then compare the mean deviation on same-family targets (e.g.\ GPT-OSS
judges $\to$ OpenAI models, Gemma judges $\to$ Google models) against the
mean deviation on cross-family targets. We repeat the same same-family minus
cross-family comparison after aggregating battles into Elo residuals. Because
LMArena 100K contains only one Qwen target, we re-run the analysis on
LMArena 140K, where Qwen coverage rises to seven and Google to ten
(Figure~\ref{fig:self-preference}).

Same-family effects are present but not universal. At the score level, the
same-family excess is small, ranging from $-0.07$ to $+0.29$ points on the
$1$--$10$ scale. At the Elo-residual level, the same-family excess is
positive on every LMArena 100K judge and on five of seven LMArena 140K judges,
with the largest values on the Gemma family in 140K (where the corpus exposes
ten Google models). Thus same-family preference can contribute to individual
residual deviations, but it is not a uniform explanation for the
strength-correlated residual ramp. Soft-Elo reduces the Elo-residual excess on
most judges but inherits any score-level excess; eliminating it would require
explicit score-stage debiasing or a multi-judge ensemble.

\subsection{Strength-correlated structure under Soft-Elo}
\label{app:residual-structure-soft}

\begin{figure}[h]
\centering
\includegraphics[width=0.7\linewidth]{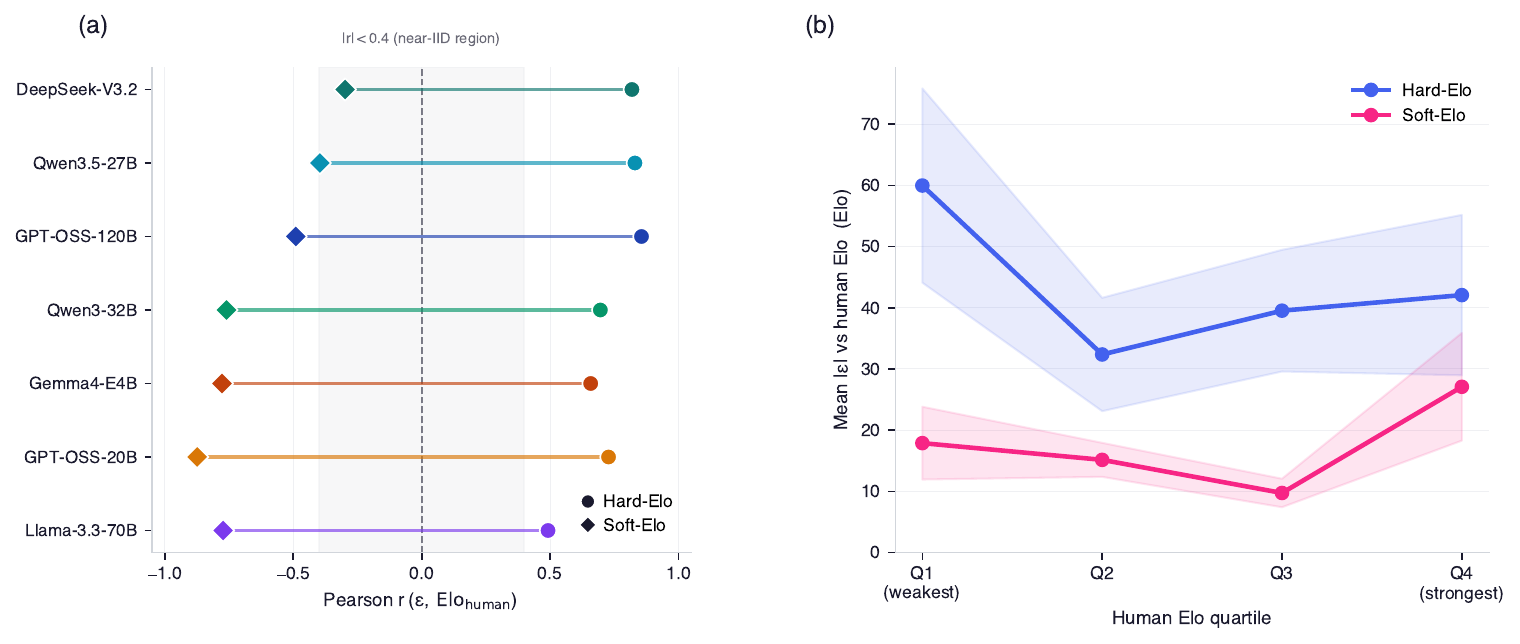}
\caption{\textbf{Strength-correlated residual structure under Soft-Elo.}
Mean signed residual $\varepsilon = \eloLLM - \eloHuman$ per human-Elo
quartile, averaged across judges (thick lines) with $\pm 1\,\mathrm{SD}$
band; faint lines per judge underneath. Hard-Elo ramps from $-61$~Elo
at Q1 to $+40$~Elo at Q4 (${\sim}100$~Elo span); Soft-Elo compresses
this to a ${\sim}36$~Elo span with the slope sign-flipped, signaling a
small uniform over-correction across all judges.}
\label{fig:residual-exchangeability}
\end{figure}

Section~\ref{sec:residual-structure-hard} showed that under Hard-Elo
the signed residual ramps positively with human Elo on every judge.
Figure~\ref{fig:residual-exchangeability} traces this ramp under both
methods using the cross-judge mean signed residual per Elo quartile,
with faint per-judge lines underneath.

Hard-Elo spans a wide ${\sim}100$~Elo ramp from Q1 to Q4 (mean
residuals $-61$, $-12$, $+36$, $+40$ across the four quartiles).
Soft-Elo compresses this ramp to ${\sim}36$~Elo ($+12$, $+8$, $+4$,
$-24$): roughly $3\times$ flatter, with the sign of the
strength--residual relationship inverted. The compression is
consistent across judges: every judge shows the same Hard slope and
the same shallow Soft anti-slope.

\paragraph{The Hard-Elo distortion is asymmetric.}
Hard-Elo's residual at Q1 ($-61$~Elo) is substantially larger in
magnitude than at Q4 ($+40$~Elo): weak models are dragged down by
roughly $50\%$ more than strong models are pushed up. A plausible
mechanism is that weak models lose the majority of their battles, so
the score-difference information Hard-Elo discards is concentrated on
the weak side; strong models win most of theirs by larger inferred
score differences regardless of the binary collapse.

\paragraph{The Soft-Elo over-correction is structured.}
Soft-Elo's residual is positive at Q1 and negative at Q4 on every
single judge. The mean residual pattern indicates a small over-correction of the strength axis. When
$\bstar$ maps score differences to preference probabilities, the
resulting BT fit pulls strong-vs-weak comparisons modestly closer to
the population mean than the human reference does. The mean over-correction is $\le 24$~Elo at the extreme quartile.

\paragraph{Frontier models remain the noisy stratum.}
Q4 carries the largest cross-judge SD under \emph{both} methods (SD
$17$ for Hard-Elo, $9$ for Soft-Elo). Soft-Elo flattens the cross-judge
mean ramp but does not eliminate per-judge variability at the strong
end; this motivates checking coverage within strength strata, alongside the
marginal coverage reported in Section~\ref{sec:discussion}.

\section{Conformal Interval Analysis}
\label{app:interval-analysis}
\subsection{Width decomposition}
\label{sec:cp-decomposition}

\begin{figure}[ht]
\centering
\includegraphics[width=0.62\linewidth]{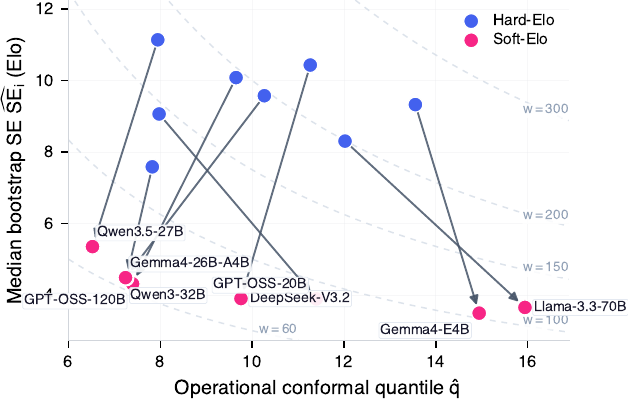}
\caption{\textbf{Width decomposition in the $(\hat q,\,\widehat{SE})$ plane.}
Each judge contributes a Hard-Elo (blue) $\to$ Soft-Elo (pink) arrow on a
background of constant-width contours $w = 2\hat q \cdot \widehat{SE}_i$.}
\label{fig:width-decomposition}
\end{figure}

The conformal interval has half-width
$\hat q \cdot \widehat{SE}_{n+1}$ (Eq.~\ref{eq:conformal-interval}), so widths
can shrink because the global conformal quantile $\hat q$ drops, because the
per-model bootstrap standard error $\widehat{SE}_i$ drops, or both.
Figure~\ref{fig:width-decomposition} plots each judge as a single summary
point at $(\hat q,\,\operatorname{median}_i \widehat{SE}_i)$, using the median
of $\widehat{SE}_i$ across held-out test models for both methods, with
arrows from the Hard-Elo position to the Soft-Elo position, on a background
of constant-width contours $w = 2\hat q \cdot \widehat{SE}$.

Every judge crosses to a lower-width contour, and the two regimes correlate
with judge--human agreement on individual battles (Table~\ref{tab:judges}):
the three $\hat q$-rises are concentrated among the lower-$\kappa$ judges
(Llama-3.3-70B, GPT-OSS-20B, Gemma4-E4B) and the $\hat q$-drops among the
higher-$\kappa$ judges, though the ordering is not strictly monotonic in
$\kappa$ (Appendix~\ref{app:residual-structure-soft} expands this analysis).
For the five down-and-left judges, both $\hat q$ and the median bootstrap
$\widehat{SE}_i$ fall together (Qwen3.5-27B representative). For the
three down-and-right judges, $\hat q$ rises modestly, but
$\widehat{SE}_i$ falls enough that the width still shrinks
(Gemma4-E4B extreme). Per-judge trajectories are in
Figure~\ref{fig:width-decomposition}. The empirical coverage band
tightens in parallel (Section~\ref{sec:cp-coverage}), so the same refit
yields narrower intervals \emph{and} empirical coverage tighter around
the nominal.

\subsection{Coverage and model assumptions}
\label{app:additional-discussion}

\paragraph{Sources of non-exchangeability.}
The conformal guarantee is marginal and relies on exchangeability between the
calibration models and the future model being evaluated. Several shifts can
challenge this assumption. Prompt-difficulty shift can occur when the calibration
set is easier than the deployment set. Model-family or post-training shift can
occur when future models come from a family or training recipe poorly represented
in the calibration pool. Temporal drift can occur when newly released models
differ systematically from the models used to calibrate the residuals. Finally,
battle-level dependence remains because the same prompt can appear across many
model pairs; our bootstrap resamples battles within each model and does not
cluster by prompt. Prompt-block bootstraps, weighted conformal
prediction~\citep{barber2023conformal}, and stratified conformal quantiles by
prompt difficulty or model family are natural extensions.

\paragraph{Relation to Bradley--Terry assumptions.}
Soft-Elo changes the targets passed into Bradley--Terry, but it does not change
the Bradley--Terry model class. The resulting leaderboard is still a scalar Elo
projection of a potentially cyclic preference relation. Preference cycles,
non-transitive comparisons, and model-pair-specific interactions therefore
remain limitations of any BT-based leaderboard. Soft-Elo addresses a different
failure mode: it prevents the BT fit from treating narrow and decisive LLM-judge
preferences as equally informative.

\clearpage
\section*{NeurIPS Paper Checklist}

\begin{enumerate}

\item {\bf Claims}
    \item[] Question: Do the main claims made in the abstract and introduction accurately reflect the paper's contributions and scope?
    \item[] Answer: \answerYes{}
    \item[] Justification: The abstract and introduction state the paper's scope:
    estimating human Elo ratings from LLM-judged battles, replacing hard
    preference targets with calibrated soft targets, and applying split
    conformal prediction to LLM--human Elo residuals. The claimed improvements
    are supported by the held-out Elo estimation and conformal interval results
    in Sections~\ref{sec:softelo} and~\ref{sec:conformal}; assumptions and
    limitations are stated in Section~\ref{sec:discussion}.

\item {\bf Limitations}
    \item[] Question: Does the paper discuss the limitations of the work performed by the authors?
    \item[] Answer: \answerYes{}
    \item[] Justification: Section~\ref{sec:discussion} discusses the main
    limitations: Soft-Elo models only the uncertainty encoded in judge score
    differences, and the conformal guarantee is marginal and relies on
    exchangeability between calibration and future models. Appendix~\ref{app:additional-discussion}
    provides additional diagnostics and caveats.

\item {\bf Theory assumptions and proofs}
    \item[] Question: For each theoretical result, does the paper provide the full set of assumptions and a complete (and correct) proof?
    \item[] Answer: \answerNA{}
    \item[] Justification: The paper does not introduce new theoretical results
    or proofs. It uses standard split conformal prediction guarantees from prior
    work, with the exchangeability assumption stated in
    Section~\ref{sec:conformal-method} and discussed as a limitation in
    Section~\ref{sec:discussion}.

    \item {\bf Experimental result reproducibility}
    \item[] Question: Does the paper fully disclose all the information needed to reproduce the main experimental results of the paper to the extent that it affects the main claims and/or conclusions of the paper (regardless of whether the code and data are provided or not)?
    \item[] Answer: \answerYes{}
    \item[] Justification: The main formulas and evaluation protocol are
    specified in Sections~\ref{sec:conformal-method} and~\ref{sec:softelo}.
    Appendix~\ref{app:judge-protocol} gives the judge prompt, score extraction,
    sampling, and corpus statistics; Appendix~\ref{app:softelo-algorithm} gives
    the full Soft-Elo pseudocode; and Section~\ref{sec:conformal} specifies the
    conformal split and bootstrap settings used for the interval results.

\item {\bf Open access to data and code}
    \item[] Question: Does the paper provide open access to the data and code, with sufficient instructions to faithfully reproduce the main experimental results, as described in supplemental material?
    \item[] Answer: \answerYes{}
    \item[] Justification: Code is available at
    \url{https://github.com/kargibora/SoftElo}. The LMArena 100K and 140K
    judge annotations and reproduction package are hosted at
    \url{https://huggingface.co/datasets/kbora/SoftElo}. Both repositories
    are public. The annotation package includes loading instructions and
    commands for reproducing the main Elo, label-smoothing, and conformal results.

\item {\bf Experimental setting/details}
    \item[] Question: Does the paper specify all the training and test details (e.g., data splits, hyperparameters, how they were chosen, type of optimizer) necessary to understand the results?
    \item[] Answer: \answerYes{}
    \item[] Justification: Sections~\ref{sec:conformal-method} and~\ref{sec:softelo}
    define the BT, Soft-Elo, and conformal procedures. Appendix~\ref{app:judge-protocol}
    gives the judge prompt, scoring rubric, decoding settings, score extraction,
    and sampling details; Appendix~\ref{app:softelo-algorithm} gives the full
    held-out evaluation pseudocode; and Section~\ref{sec:conformal} specifies
    the conformal split sizes, miscoverage level, and bootstrap resamples.

\item {\bf Experiment statistical significance}
    \item[] Question: Does the paper report error bars suitably and correctly defined or other appropriate information about the statistical significance of the experiments?
    \item[] Answer: \answerYes{}
    \item[] Justification: The conformal interval experiments report empirical
    coverage ranges across random model-level splits and use bootstrap standard
    errors in the nonconformity scores (Section~\ref{sec:conformal}). Figures
    involving binned agreement or residual summaries state the uncertainty bands
    or error bars in their captions, and Appendix~\ref{sec:cp-decomposition}
    decomposes interval-width changes into conformal-quantile and bootstrap-scale
    effects.

\item {\bf Experiments compute resources}
    \item[] Question: For each experiment, does the paper provide sufficient information on the computer resources (type of compute workers, memory, time of execution) needed to reproduce the experiments?
    \item[] Answer: \answerYes{}
    \item[] Justification: Appendix~\ref{app:compute-resources} reports the
    compute environment used for judge annotation and local post-processing.
    Open-weight judges were run on NVIDIA A100-SXM 64GB GPUs with 1--4 GPUs
    per job depending on model size, DeepSeek-V3.2 was run through the OpenRouter
    API, and the remaining BT, bootstrap, conformal, and plotting steps
    were CPU-based post-processing on an Apple-silicon laptop over cached judge
    outputs.

\item {\bf Code of ethics}
    \item[] Question: Does the research conducted in the paper conform, in every respect, with the NeurIPS Code of Ethics \url{https://neurips.cc/public/EthicsGuidelines}?
    \item[] Answer: \answerYes{}
    \item[] Justification: The main study uses public benchmark datasets and
    LLM-generated annotations. The additional 200-battle re-annotation was
    performed independently by one of the authors; no participants were
    recruited. Limitations and responsible use are discussed in
    Section~\ref{sec:discussion} and Appendix~\ref{app:additional-discussion}.
    The absence of formal ethics review is disclosed in item 15.

\item {\bf Broader impacts}
    \item[] Question: Does the paper discuss both potential positive societal impacts and negative societal impacts of the work performed?
    \item[] Answer: \answerYes{}
    \item[] Justification: Section~\ref{sec:discussion} includes a broader
    impact statement: the positive impact is lower-cost model evaluation with
    calibrated model-level uncertainty, while the main negative risk is
    over-trusting LLM-judge leaderboards or conformal intervals when judge
    biases or exchangeability violations remain unaddressed. The introduction
    motivates the evaluation setting, and Appendix~\ref{app:additional-discussion}
    gives additional caveats on prompt, model-family, and temporal shift.

\item {\bf Safeguards}
    \item[] Question: Does the paper describe safeguards that have been put in place for responsible release of data or models that have a high risk for misuse (e.g., pre-trained language models, image generators, or scraped datasets)?
    \item[] Answer: \answerNA{}
    \item[] Justification: The paper does not release a new pretrained model,
    image generator, or scraped dataset with high misuse risk. The released
    assets are anonymized code/configuration files and derived evaluation
    artifacts; the source datasets are existing public benchmark datasets cited
    where they are introduced.

\item {\bf Licenses for existing assets}
    \item[] Question: Are the creators or original owners of assets (e.g., code, data, models), used in the paper, properly credited and are the license and terms of use explicitly mentioned and properly respected?
    \item[] Answer: \answerYes{}
    \item[] Justification: We cite the source papers and dataset cards for
    LMArena, Chatbot Arena, ComparIA, and the judge models used in the study.
    Appendix~\ref{app:licenses} consolidates the licenses of all datasets
    (LMArena prompts under CC-BY-4.0 with model outputs governed by provider
    terms; ComparIA votes under Etalab 2.0) and all eight judge models (six
    under Apache 2.0, one under MIT, and Llama-3.3-70B-Instruct under the
    Llama 3.3 Community License Agreement). The released annotation files
    exclude raw prompts and model completions, and judge weights are accessed
    through their publicly hosted releases under the cited terms.

\item {\bf New assets}
    \item[] Question: Are new assets introduced in the paper well documented and is the documentation provided alongside the assets?
    \item[] Answer: \answerYes{}
    \item[] Justification: The code repository at
    \url{https://github.com/kargibora/SoftElo} includes usage instructions.
    The annotation package at
    \url{https://huggingface.co/datasets/kbora/SoftElo} includes a dataset card,
    field definitions, source attribution, licenses, and reproduction commands.
    The appendix documents the corpora, judge prompt, generation settings,
    score extraction, and sampling procedure.

\item {\bf Crowdsourcing and research with human subjects}
    \item[] Question: For crowdsourcing experiments and research with human subjects, does the paper include the full text of instructions given to participants and screenshots, if applicable, as well as details about compensation (if any)? 
    \item[] Answer: \answerNA{}
    \item[] Justification: The main experiments reuse public human-preference
    datasets. One of the authors independently re-annotated 200 LMArena
    battles for the agreement analysis reported in the rebuttal. No
    participants were recruited and no crowdsourcing was used.

\item {\bf Institutional review board (IRB) approvals or equivalent for research with human subjects}
    \item[] Question: Does the paper describe potential risks incurred by study participants, whether such risks were disclosed to the subjects, and whether Institutional Review Board (IRB) approvals (or an equivalent approval/review based on the requirements of your country or institution) were obtained?
    \item[] Answer: \answerNo{}
    \item[] Justification: No formal institutional ethics review or IRB
    approval was obtained. The study reuses public benchmark data, and the
    additional re-annotation was performed independently by one of the
    authors without recruiting participants.

\item {\bf Declaration of LLM usage}
    \item[] Question: Does the paper describe the usage of LLMs if it is an important, original, or non-standard component of the core methods in this research? Note that if the LLM is used only for writing, editing, or formatting purposes and does \emph{not} impact the core methodology, scientific rigor, or originality of the research, declaration is not required.
    \item[] Answer: \answerYes{}
    \item[] Justification: LLMs are a core component of the research because
    they serve as judge models whose scores are calibrated and compared to human
    Elo ratings. The paper describes the judge protocol, generation settings,
    score extraction, and evaluated judge pool in Sections~\ref{sec:background}
    and~\ref{sec:softelo} and Appendix~\ref{app:judge-protocol}.

\end{enumerate}

\end{document}